\documentclass[ijoc,nonblindrev]{informs3} 
\usepackage[utf8]{inputenc}
\usepackage[english]{babel}
  \usepackage{subcaption}
  \usepackage{gensymb}
\usepackage{microtype}
\usepackage{graphicx}
\usepackage{booktabs} 

\usepackage{microtype}
\usepackage{booktabs} 
\usepackage{hyperref}
\usepackage{amsmath,dsfont,amsfonts}
\usepackage{pifont,textcomp,bbm}
\usepackage{algorithm, algpseudocode}
 \usepackage{multirow} 

\OneAndAHalfSpacedXII 

\usepackage{natbib}
 \bibpunct[, ]{(}{)}{,}{a}{}{,}%
 \def\bibfont{\small}%
\TheoremsNumberedThrough     

\EquationsNumberedThrough    
\newcommand\independent{\protect\mathpalette{\protect\independenT}{\perp}}
\def\independenT#1#2{\mathrel{\rlap{$#1#2$}\mkern2mu{#1#2}}}

\begin{document}


\RUNAUTHOR{Wang and Rudin}

\RUNTITLE{Causal Rule Sets}

\TITLE{Causal Rule Sets for Identifying Subgroups with Enhanced Treatment Effects}

\ARTICLEAUTHORS{%
\AUTHOR{Tong Wang}
\AFF{Tippie College of Business}
\AFF{University of Iowa}
\AFF{\EMAIL{tong-wang@uiowa.edu}}
\AUTHOR{Cynthia Rudin}
\AFF{Department of Computer Science}
\AFF{Duke University}
\AFF{\EMAIL{cynthia@cs.duke.edu}}
} 

\ABSTRACT{%
 A key question in causal inference analyses is how to find subgroups with elevated treatment effects. This paper takes a machine learning approach and introduces a generative model, Causal Rule Sets (CRS), for interpretable subgroup discovery. A CRS model uses a small set of short decision rules to capture a subgroup where the average treatment effect is elevated. We present a Bayesian framework for learning a causal rule set. The Bayesian model consists of a prior that favors simple models for better interpretability as well as avoiding overfitting, and a Bayesian logistic regression that captures the likelihood of data, characterizing the relation between outcomes, attributes, and subgroup membership. The Bayesian model has tunable parameters that can characterize subgroups with various sizes, providing users with more flexible choices of models from the \emph{treatment efficient frontier}. We find maximum \textit{a posteriori} models using iterative discrete Monte Carlo steps in the joint solution space of rules sets and parameters. To improve search efficiency, we provide theoretically grounded heuristics and bounding strategies to prune and confine the search space. Experiments show that the search algorithm can efficiently recover true underlying subgroups. We apply CRS on public and real-world datasets from domains where interpretability is indispensable. We compare CRS with state-of-the-art rule-based subgroup discovery models. 
 Results show that CRS achieved consistently competitive performance on datasets from various domains, represented by high treatment efficient frontiers.
}%



\maketitle
\section{Introduction}
When estimating the efficacy of a treatment from observational data, we may find that the treatment is effective only for a subgroup of the population.
One often-cited example in oncology is the drug trastuzumab, which has been shown to be an effective treatment for breast cancer only when the tumor is HER2 positive \citep{baselga2006adjuvant}. This example is illustrative of the necessity to identify such subgroups if they exist.  Similar examples can be found in personalized therapies, targeted advertising, and other types of personalized recommendations.  
Our goal in this paper is to discover such subgroups from observational data, where a treatment has an elevated effect with respect to the rest of the population. 

Problems related to this subject are referred to as \emph{subgroup analysis} in the context of causal inference \citep{rothwell2005subgroup}.  Prior research can be classified into two categories.
One category of methods, called moderated regression analysis \citep{cohen2013applied}, fits statistical regression models that include possible treatment-feature interactions. The methods in this category mainly test the hypotheses of known subgroups instead of discovering unknown subgroups from data; these methods rely on clear prior hypotheses about which subgroups could be involved in the interactions. However, such \textit{a priori} hypotheses may not exist in many applications. Particularly when the number of features is large, it becomes harder for domain experts to identify which of myriad possible subgroups to consider.
The other category of widely adopted approaches uses tree-based recursive partitioning \citep{su2009subgroup,lipkovich_subgroup_2011} to partition data into smaller subsets until a stopping criterion is met. 
Recursive partitioning methods are a natural way to analyze a large number of features that have potentially complicated interactions. 
But they partition data greedily using heuristics, and the splits are not based on the treatment effect further down in the tree. There exist several issues with greedily learned trees, as discussed in many recent works \citep{brijain2014survey,kim2016hybrid}. Specifically, the downfall of being greedy in growing a tree is that there is no direct optimization of a global objective, and it often yields suboptimal solutions.

In this work, we learn subgroups directly from observational data  under a carefully designed global objective -- without a prespecified hypothesis of what subgroups to consider. Our method, Causal Rule Sets (CRS), captures a subgroup with a small set of short decision  rules, where each rule is a conjunction of conditions. The rules and the number of rules are learned from data. An instance is in the subgroup if it satisfies at least one rule in the set.  An example of a CRS model is shown below. The treatment is giving a coupon to a potential customer, and the outcome is whether the customer buys a product.

  \noindent\textbf{if} customer (is older than 40 \emph{AND} has children \emph{AND} annual income $<$\$50,000) \\  \emph{OR} (is female \emph{AND} married \emph{AND} used the coupon before) \\
 \textbf{then} the customer is more highly influenced by the coupon.

In this example, the model consists of two rules. It captures customers who are more likely to be influenced by a coupon compared to the general population of customers. 

\begin{figure}[ht]
\centering
\includegraphics[width=0.6\textwidth]{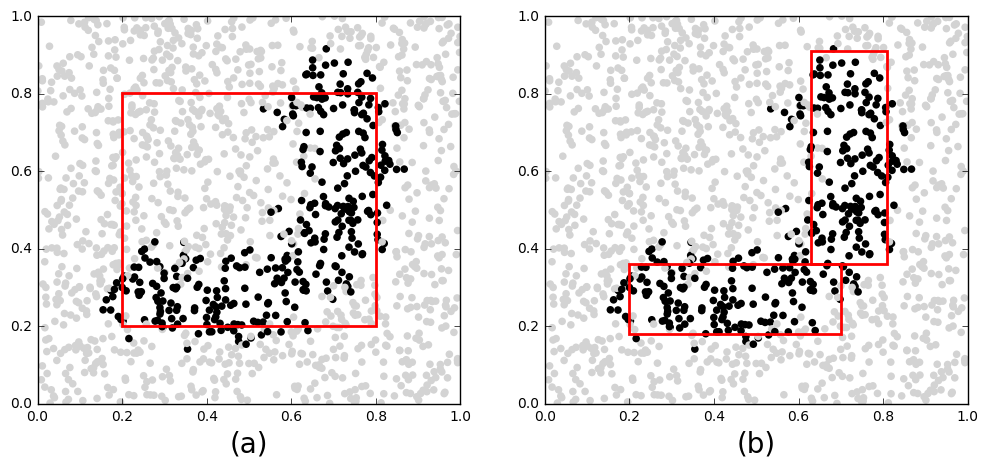}
\caption{An example of a subgroup. Black dots represent a true subgroup, and gray dots represent the rest of the population. (a) corresponds to one rule. (b) corresponds to a rule set of two rules.}\label{fig:demo}
\end{figure}

Unlike tree-based greedy methods that partition data into different regions and define each group with one rule, our model uses a set of (non-greedy) rules to capture a subgroup. We illustrate this distinction using the example below.
Imagine data distributed on a 2D plane. Using one rule places an implicit constraint on identified subgroups: they are limited to points that fit into one rectangle. For example,  ``$0.2\leq X_1 \leq 0.4 \emph{ AND } 0.3\leq X_2 \leq 0.6$'' is represented by the red rectangle in Figure~\ref{fig:demo}(a). If the subgroup has other irregular shapes, for example, distributed along an arc, as represented by black dots in Figure~\ref{fig:demo}(a), it cannot be captured by a single rule.  
However, it can be identified via multiple rules. It is easy to find a set of smaller rectangles that collectively cover the subgroup, as shown in Figure 1(b).  As the shape becomes more irregular, it becomes a question of how to place a set of rectangles to cover the points. Allowing more rectangles will cover the points more accurately but at the cost of model simplicity, while using few rectangles may lose the precision of capturing the true subgroup. Our method does not use greedy recursive partitioning but optimizes a global objective that considers both the quality of the fit to the data and the complexity of the model (the number of rules).

A rule set brings greater freedom in defining a subgroup and can potentially uncover more complicated interactions among features. Rule sets are also referred to as disjunctive normal form models (DNF) \citep{cook1971complexity}.
Subgroups captured by DNF are not new in the literature \citep{del2007evolutionary, atzmueller2006sd,novak2009supervised,kavvsek2006apriori,carmona2009analysis,wang2017bayesian}. Prior research developed various rule induction algorithms to find a set of rules to capture an interesting, or unusual subgroup, with regard to a target variable.
However, their goals are fundamentally different from ours in this paper: previous works discover subgroups in the context of classification, where there are known target variables in the training set, and the objective considers minimizing the discrepancy between the predicted outcomes and the real outcomes. Our problem is in the context of causal inference. The optimizing objective involves the treatment effect, which is the difference in the potential outcomes. Classification is supervised, whereas for causal inference, it is half-supervised; we have only half of each label (either the treatment or the control outcomes but not both), and the other half needs to be inferred.

In this work, we propose an efficient inference method for learning a maximum \emph{a priori} (MAP) model. The algorithm consists of two steps to effectively reduce computation. First, a rule space is constructed via rule mining.
Then, the search algorithm iterates over states that consist of rule sets and the corresponding parameters until convergence. This algorithm applies an exploration-with-exploitation strategy to improve search efficiency. We also propose search criteria that make it easier to efficiently find promising solutions. 

We conducted experiments to test the ability of CRS to retrieve true subgroups. 
Our Bayesian framework can generate rule sets of various sizes, produced with different hyperparameters. To evaluate the performance, we propose a novel metric, the \textbf{treatment efficient frontier}, which characterizes how quickly the average treatment effect on the subgroup decays as the support of the subgroup (the amount of test data in the subgroup) grows.  We compared CRS models with other baseline models using real-world data sets. 
Compared to baseline methods, CRS demonstrated competitive out-of-sample performance while offering greater flexibility of obtaining subgroups with various support and average treatment effect, as shown by the higher treatment efficient frontiers.

 \section{Related Work}
 Below, we discuss related work on subgroup discovery for causal inference and rule-based models.

    \subsection{Subgroup discovery in observational causal inference}\label{sec:related}
     Subgroup discovery methods (also called pattern detection in some contexts) are used to identify groups of units that are somehow ``interesting,'' according to a measure of interestingness \citep{lemmerich2016fast}. In our case, an interesting subgroup is one with elevated estimated treatment effect.   
 There is an extremely large body of work on subgroup discovery for non-causal problems \citep{atzmueller2015subgroup,gamberger2002expert,lavrac2004subgroup,atzmueller2006sd}.
 We will review several subfields of subgroup discovery that pertain to observational causal inference.
 


   
\textit{Subgroup discovery with a small number of pre-defined hypotheses.}  
    A classic type of method pertains to the situation where clear \emph{a priori} hypotheses exist about which subgroups have elevated or reduced treatment effect. Then statistical methods are applied to assess the hypotheses \citep{assmann2000subgroup, lagakos2006challenge}. For example,
    \cite{malhotra2019effects} and \cite{solomon2018effect} study the treatment effect on a subgroup defined by different factors such as a certain genotype. 
    
\textit{Subgroup analysis using post-hoc analysis on estimated potential outcomes.} Since observational causal inference problems do not have both a treatment outcome and a control outcome measured for each unit, recent works model these potential outcomes and perform posthoc analysis on them (along with the observed outcomes) to find interesting subgroups. 
    One approach is to first perform matching using a high-quality matching method \citep[e.g.,][]{FLAME}, and then use a subgroup discovery or classification method on the matched groups. Similarly, Virtual Twins (VT) \citep{foster_subgroup_2011} first uses two random forests to estimate the probability of positive outcomes in treatment and control groups, respectively, and then uses the difference of the two as a target variable to build a regression or classification tree. However, creating the rules directly from the data is more straightforward and does not rely on the quality of the extra step of matching.
    
\textit{Directly discovering subgroups from data.} A third type of method directly finds subgroups from data without a \emph{priori} hypotheses or estimating the potential outcomes.
A few recent and widely cited methods include:  
Subgroup Identification based on Differential Effect Search (SIDES) \citep{lipkovich_subgroup_2011} and Qualitative Interaction Trees (QUINT) \citep{dusseldorp2014qualitative}, which are tree-based recursive partitioning methods, and they differ mainly in their partitioning criteria. The goal is to construct trees such that their leaves cover instances with elevated treatment effect.  SIDES and QUINT are used as baselines in this paper. They have a disadvantage in that they are greedy methods that are not designed to attain global optima of any objective function. \cite{mcfowland2018efficient} proposed Treatment Effect Subset Scan, a method motivated by pattern detection methods  \citep{neill2012fast} that maximizes a nonparametric scan statistic over subpopulations, but the method is designed for randomized data are does not apply to observational data.
Recent work \citep{nagpal2020interpretable} proposes an interpretable model, called the Heterogeneous Effect Mixture Model (HEMM), to discover subgroups with enhanced or diminished causal effect due to treatment.  HEMM uses  mixture distributions rather than rule sets to capture subgroups, thus yielding soft assignments to subgroups. The work of \cite{MorucciEtAl20} finds a hyperbox of constant treatment effect around any given unit, but is not designed to find elevated treatment effects in particular. Another work similar to ours also uses a set of rules, i.e., Causal Rule Ensemble, to capture a subgroup \citep{lee2020causal}. CRE uses a linear combination of rules, where each rule in the linear combination is a leaf of a decision tree. The authors create an ensemble of trees and any leaf in any tree can be used for the combination. The rules themselves are not optimized for sparsity, and the linear combination is optimized for sparsity using an $\ell_1$ regularization parameter. 
 
Our CRS falls into this third category of methods. Similar to SIDES and QUINT, CRS is also rule-based. But unlike the tree structures, our model is comprised of a set of association rules, also called Disjunctive Normal Form models (DNF), to capture a subgroup.  
DNFs have been used to describe subgroups of interest \citep{del2007evolutionary, atzmueller2006sd,novak2009supervised,kavvsek2006apriori,carmona2009analysis}, and in this paper, we apply them to observational causal inference. 

 \subsection{Rule-based Interpretable Models}
 Rule-based models are popular forms of interpretable models since decision rules are easy to understand because of their symbolic forms and intuitive logical structures. 
 There have been a number of recent works on rule-based interpretable models, including rule sets (unordered rules) \citep{wang2017bayesian, rijnbeek2010finding,wang2019hybrid}, rule lists (ordered rules) \citep{ynormalize_addang2016scalable,angelino2017learning}, or rules assembled into other logic forms \citep{dash2018boolean,pmlr-v108-pan20a}.
 These models are classifiers, and they were not conceived to model treatment effects. Recently there have been works on finding treatment regimes \citep{moodie2012q, lakkaraju2017learning} using rule lists. Their goal is to stratify data into different subspaces with various treatment effects, whereas our goal is to find one subgroup with an enhanced treatment effect compared to that of the entire population. One could view the problem of finding a subgroup with elevated treatment as a simplified problem of finding an optimal treatment regime.
 
 Learning a rule-based model is a challenging task. Existing algorithms for learning rule-based models either use sampling or optimization approaches to create models \citep{letham2015interpretable, wang2018multi,wang2017bayesian,lakkaraju2016interpretable,angelino2017learning,dash2018boolean,wei2019generalized}.
 Our CRS model is based on simulated annealing, and we customize the proposal strategies, incorporating a theoretical understanding of CRS models to improve search efficiency.

\section{Causal Rule Sets}
We work with data $\mathcal{D} = \{(\mathbf{x}_i,y_i,T_i)\}_{i=1}^n$. Each instance is described by a covariate vector $\mathbf{x}_i\in \mathbb{R}^{J+1}$ that consists of $J$ features and a constant to account for an intercept term, and each $\mathbf{x}_i$ is drawn iid from a distribution over $\mathbf{x}$.  
Let $T_i\in\{0,1\}$ denote the treatment assignment for instance $i$ and $T_i = 1$ represents receiving the treatment. $T_i$ is assumed to be randomly assigned, possibly depending on $\mathbf{x}_i$.
The outcome is represented by $y_i\in\{0,1\}$ and we assume $y_i = 1$ is the desired outcome. 
We use the potential outcomes framework \citep{rubin1974estimating} with potential outcomes $Y_i(1), Y_i(0) \in \{0,1\}$ under treatment and control. The outcomes are drawn independently from conditional distributions $Y(1)|\mathbf{x}$ or $Y(0)|\mathbf{x}$. The treatment effect at $\mathbf{x}$ is defined as
\begin{equation}
\tau(\mathbf{x}) = \mathbb{E}\big[ Y(1) - Y(0)|\mathbf{x}\big].
\end{equation}
We maintain the following canonical assumptions of observational inference \citep{angrist1996identification,MorucciEtAl20}:
\begin{enumerate}
    \item \textbf{Overlap}: For all values of $\mathbf{x}$ for instance $i$, $0< \text{Pr}(T_i = 1|\mathbf{X}_i = \mathbf{x})<1$.
    \item \textbf{SUTVA}: Stable Unit Treatment Values Assumption, which refers to (i) the potential outcomes for any unit do not vary with the treatments assigned to other units (no interference), and (ii)
for each unit, there are no different forms or versions of each treatment level, which lead to different potential outcomes (no hidden variations of treatment).
    \item \textbf{Conditional ignorability}: For all instances and any $t \in \{0,1\}$, treatment is administered independently of outcomes conditionally on the observed covariates, i.e., $T_i \independent Y_i(1), Y_i(0)|\mathbf{X}_i = \mathbf{x}_i$. This directly implies that $\mathbbm{E}[Y_i|T = t, \mathbf{X}_i = \mathbf{x}_i] = \mathbbm{E}[Y_i(t)|\mathbf{X}_i = \mathbf{x}_i],$ which enables us to estimate treatment effects using the observed data.
\end{enumerate}

Given a dataset $\mathcal{D}$, our goal is to find a set of rules to capture a subgroup that demonstrates an enhanced average treatment effect. 
The main difficulty in finding such a subgroup is one can only observe one of the two potential outcomes for a given instance. Therefore one cannot directly estimate the differences in the form $y_i(1) - y_i(0)$ at the individual level. This is also common to most problems in causal analysis. 
Under the above assumptions, however, we may be able to treat nearby observations as similar to those that may have arisen from a randomized experiment. 
However, we need to know which observations can be considered ``similar'' because the similarity in features does not necessarily translate into similarity in treatment effects. The training set can help determine which units are similar in terms of potential outcomes and can thus be used to estimate treatment effects, which in turn,  determines which units to include within our subgroup. 

A rule is a conjunction of conditions. For example, ``female \emph{AND} age $>$ 50 \emph{AND} married'' has three conditions. Let $a(\mathbf{x}_i)\in\{0,1\}$ represent whether $\mathbf{x}_i$ satisfies rule $a$, i.e., $\mathbf{x}_i$ is ``\emph{covered}'' by rule $a$.
Let $A$ denote a set of rules. An observation satisfies the rule set $A$ if it satisfies at least one rule in $A$, represented as below. We abuse the notation of $A(\cdot)$ to represent whether an instance belongs to the subgroup defined by $A$:
\begin{equation}\label{eqn:defA}
A(\mathbf{x}_i)=  \begin{cases} 1 & \exists a \in A: a(\mathbf{x}_i)=1 \\  0 & \text{otherwise.} \end{cases}
\end{equation}
The goal is to find $A$ to cover a subgroup where the treatment demonstrates an enhanced treatment effect.  Let us provide some definitions.
\begin{definition}
$A$ is a set of association rules and (using overloaded notation) function $A(\cdot)$ is defined in (\ref{eqn:defA}).
Given a dataset $\mathcal{D}$, $A$ is a \textit{causal rule set} on $\mathcal{D}$ if $$
    \mathbb{E}\big[ Y(1) - Y(0)|\mathbf{x}:A(\mathbf{x})=1] >     \mathbb{E}\big[ Y(1) - Y(0)]. $$
\end{definition}
\begin{definition}
The \textit{support} of a subgroup captured by rule set $A$ is the number of training instances that satisfy at least one rule in $A$, i.e., $\text{supp}(A) = \sum_{i}^n A(\mathbf{x}_i)$.
\end{definition}

In practice, a more reliable subgroup would be a large subgroup with large treatment effect. For a collection of causal rule sets, there  exists a frontier of causal rule set models, where none of them is dominated by the others in both metrics; here, a model dominates another if it is larger in both support and average treatment effect than the other. We call the collection of non-dominated models the ``treatment efficient frontier,'' which will be discussed and studied in detail in the experiments section. 

We propose a generative framework for learning a causal rule set $A$. The generative approach turns this problem into finding a MAP solution $P(A|\mathcal{D},H)$, given data $\mathcal{D}$ and a set of hyperparameters denoted by $H$ which will be discussed in detail later. The model consists of a prior $\text{Prior}(A;H)$ for complexity control (model regularization) and a Bayesian logistic regression $\Theta(\mathcal{D};A, H)$ for modeling the conditional likelihood of data, considering the relation between multivariate features, treatment, and subgroup membership.

\subsection{Prior}
We use a generative process described in \cite{wang2017bayesian} for the prior that favors a simple model, i.e., a \emph{small} set of \emph{short} rules. A small model is easy to interpret since it contains fewer conditions to comprehend, as demonstrated by the human evaluation in recent work on interpretable rule-based models \citep{wang2018multi}. Meanwhile, shorter rules often have larger support since there are fewer conditions to satisfy, which naturally avoids overfitting.  
Let us provide some basic definitions below.
\begin{definition}
The number of conditions in a rule is called the \textit{length} of the rule.
\end{definition}
Let $\mathcal{A}$ denote a rule space partitioned into subspaces of pre-mined candidate rules by their lengths, i.e.,  $\mathcal{A} = \mathcal{A}_1\cup\cdots\mathcal{A}_L$, $L$ being the maximum rule length a user allows. $\mathcal{A}_l$ represents a set of candidate rules with length $l$. We generate a causal rule set $A$ by drawing rules from $\{\mathcal{A}_l\}_{l=1}^L$. Each rule from $\mathcal{A}_l$ is independently selected for inclusion in the rule set with probability $p_l$, which is drawn from a beta prior.
More specifically, from $\mathcal{A}$, we will choose a set of rules with different lengths, selected from corresponding candidate sets. We may choose, for instance, one rule of length 1 from $\mathcal{A}_1$, 0 rules of length 2 from $\mathcal{A}_2$, 4 rules of length 3 from $\mathcal{A}_3$ and no other rules from any of the other $\mathcal{A}_l$'s.
\begin{align}
\text{selecting a rule} \in\mathcal{A}_l \sim \text{Bernoulli}(p_l),\\
 p_l \sim\text{Beta}(\alpha_l,\beta_l).
\end{align}
Smaller $\frac{\alpha_l}{\beta_l}$ indicates that rules in pool $l$ have a smaller chance to be selected into $A$. Thus we usually choose $\alpha_l \ll \beta_l$.  $\{\alpha_l,\beta_l\}_{l=1}^L$ jointly control the expected number of rules in $A$. 
Let $M_l$ notate the number of rules drawn from pool $\mathcal{A}_l$. Then the prior for $A$ is a product of probabilities of selecting rules from each pool, i.e.,  
\begin{equation}\label{eqn:prior}
\text{Prior}(A;\{\alpha_l,\beta_l\}_{l=1}^L) = \prod_{l=1}^L p(M_l;\alpha_l,\beta_l) 
\propto \prod_{l=1}^L  B(M_{l}+\alpha_l, |\mathcal{A}_{l}| - M_{l}+\beta_l).
\end{equation}
$B(\cdot)$ represents the Beta function. Equation (\ref{eqn:prior})  is derived by integrating the Bernoulli probabilities over the distribution of the beta prior, yielding a Beta-Binomial distribution \citep{wang2017bayesian}.

\subsection{Conditional Likelihood}
Given a constructed causal rule set $A$, we formulate the conditional likelihood of data given $A$. We use a Bayesian logistic regression model to 1) control confounding covariates and 2) characterize the effect of receiving the treatment and being in the subgroup, which yields: 
\begin{equation}\label{eqn:llh}
p(Y_i=1|\mathbf{x}_i,T_i) = \sigma\Big(\mathbf{v}\mathbf{x}_i + \gamma^{(0)} T_i + \gamma^{(1)} A(\mathbf{x}_i) + \gamma^{(2)} T_iA(\mathbf{x}_i)\Big),\end{equation}
where $\sigma(\cdot)$ is a sigmoid function. Here, $p(Y=1|\mathbf{x},T)$ models the potential outcome for a unit with feature vector $\mathbf{x}$ and treatment $T$. The estimated treatment effect for a unit would be $p(Y=1|\mathbf{x},T=1) - p(Y=1|\mathbf{x},T=0)$.
The upper index in formula (\ref{eqn:llh}) models the interaction between the attributes, treatment assignment $T_i$, and the rule set $A$. $\mathbf{v}$ is a vector of coefficients for attributes including an intercept.   $\mathbf{v}\mathbf{x}_i$ captures the baseline contribution from the attributes towards receiving outcome $y_i=1$, regardless of whether receiving a treatment. (If desired, the linear baseline model can be replaced with nonlinear baseline terms that are functions of $\mathbf{x}_i$.) $\gamma^{(0)}T_i$ captures the baseline treatment effect that applies to the whole population, regardless of whether they belong to the subgroup or not.  $\gamma^{(1)}A(\mathbf{x}_i)$ models the potential baseline effect of being in subgroup $A$. This term is created to account for subgroups whose (elevated) treatment outcomes are correlated or anticorrelated with their control outcomes. Finally, $\gamma^{(2)} T_i A(\mathbf{x}_i)$ represents the additional treatment effect for subgroup $A$ (which has an elevated treatment effect). 

Having a linear model within the sigmoid is a classic and widely adopted way to model potential outcomes in causal analysis \citep{cleophas2002subgroup,chetty2016effects,figlio2014effects}. \textit{The difference is that in previous works, the subgroups are first hypothesized by domain knowledge and then evaluated using regression, whereas in this work, $A$ is unknown and our goal is to use a generative framework to find the most probable subgroup.}

As usual, we transform the linear model into a likelihood via a logistic function to represent the model's fitness to the data.
Let $\mathbf{w} = \{\mathbf{v},\mathbf{\gamma}^{(0)},\gamma^{(1)},\gamma^{(2)}\}$. Assuming data is generated iid, the conditional likelihood of data $\mathcal{D}$ is\begin{equation}\label{eqn:cllh}
\mathcal{L}(\mathcal{D};A,\mathbf{w}) = \prod_{i=1}^n p\big(y_i = 1|\mathbf{x}_i, T_i\big)^{y_i}\cdot p\big(y_i = 0|\mathbf{x}_i, T_i\big)^{(1-y_i) },
\end{equation}
parameterized by:
\begin{equation}
A \sim \text{Prior}(A;\{\alpha_l,\beta_l\}_{l=1}^L)\text{ and }
\mathbf{w} \sim \mathcal{N}(\mathbf{\mu},\Sigma).
\end{equation}
$\mathcal{N}(\mathbf{\mu},\Sigma)$ represents a normal distribution with mean $\mu$ and variance matrix $\Sigma$.
We define a partial posterior that does not include the prior information for $A$:
\begin{equation}\label{eqn:theta}
\log\Theta(\mathcal{D};A,\mathbf{w}) :=\log \mathcal{L}(\mathcal{D};A,\mathbf{w}) + \log p(\mathbf{w}),
\end{equation}
which considers both the fitness of rule set $A$ to the data  (conditional likelihood) and the regularization. If we set $\mu$ to 0, then the above form is reduced to logistic regression with $l_2$ regularization.

We cast this formulation as a machine learning problem, which means our goal is to obtain a maximum \emph{a posteriori} CRS model that performs well with respect to the likelihood on data drawn from the population distribution $p(\mathbf{x})$ (out of sample). 
We write out the learning objective as
\begin{equation}
  F(A, \mathbf{w};\mathcal{D},H) =\log\Theta(\mathcal{D};A,\mathbf{w}) + \log \text{Prior}(A; H),
\end{equation}
where $\textrm{Prior}(A;H)$ is defined in Equation (\ref{eqn:prior}). $H$ denotes a set of hyperparameters:
$H = \{\mathcal{A},L,\{\alpha_l,\beta_l\}_{l=1}^L,\Sigma,\mu\}$.
 We write the objective as $F(A,\mathbf{w})$, the Bayesian logistic regression partial posterior as $\Theta(A,\mathbf{w})$ and the prior for $A$ as $ \text{Prior}(A)$, omitting dependence on hyperparameters $H$ and data $\mathcal{D}$  when appropriate.
\section{Model Fitting}\label{sec:ModelFitting}
In this section, we describe a search procedure to find the maximum \emph{a posteriori} $(A^*, \mathbf{w}_{A^*})$ such that: 
\begin{equation}A^*, \mathbf{w}_{A^*} \in \textrm{arg}\max_{A,\mathbf{w}} F(A,\mathbf{w}).\end{equation}

    The main search procedure follows the steps of simulated annealing, which generates a Markov chain that starts from a random state, i.e., a randomly generated rule set, and proposes the next state by selecting from its neighbors. Each state is defined as $s^{[t]} = (A^{[t]},\mathbf{w}_{A^{[t]}})$, indexed by the time stamp $t$. We define the neighbors of a rule set $A^{[t]}$ to be rule sets whose edit distance is 1 to $A^{[t]}$ (one of the rules in $A^{[t+1]}$ is different from $A^{[t]}$). Therefore, $A^{[t+1]}$ is generated by \textbf{adding}, \textbf{cutting} or \textbf{replacing} a rule from $A^{[t]}$.  Then, given rule set $A^{[t+1]}$, the corresponding optimal parameters $\mathbf{w}_{A^{[t+1]}}$ can be obtained  by maximizing $\Theta(A;\mathbf{w})$ (see (\ref{eqn:theta})) since only $\Theta(A;\mathbf{w})$ in $F(A,\mathbf{w})$ depends on $\mathbf{w}$.  
 \begin{equation}\label{eqn:wa}
 \mathbf{w}_{A^{[t+1]}} \in \textrm{arg} \max_{\mathbf{w}}\Theta(A^{[t+1]},\mathbf{w}).  
 \end{equation}
 Next, with a temperature schedule function over a total of $N_\text{iter}$ steps, $B(t) = B_0^{1 - \frac{t}{N_\text{iter}}}$, the proposed state $(A^{[t+1]},\mathbf{w}_{A^{[t+1]}})$ is accepted as $s^{[t+1]}$ with probability  $\max\left\{1,\exp\left(\frac{F(A^{[t+1]},\mathbf{w}_{A^{[t+1]}})-F(s^{[t]})}{B(t)}\right)\right\}$. The Markov search chain converges as the temperature cools.

We notice that if strictly following simulated annealing steps, the algorithm will face a significant computational challenge, as discussed in previous works \citep{wang2017bayesian}. The model space grows exponentially with the number of rules, and the number of rules grows exponentially with the number of conditions, generating an immense search space. Meanwhile, at each step, there are too many neighbors to choose from. Classic simulated annealing that randomly proposes a neighbor converges very slowly, taking tens of thousands of iterations to converge, according to \citet{letham2015interpretable,WangRu15}, for even a medium-sized dataset. 
These factors make the problem difficult to solve in practice. 

To address the problem, we incorporate customized strategies to improve search efficiency, utilizing the theoretical properties and the specific structure of our model. First, we generate a set of candidate rules $\mathcal{A}$ that contains promising rules with non-negligible support and restrict our search within this set. Then we replace the random proposing with an exploitation-with-exploration strategy and use theoretically grounded heuristics to reduce computation.  We describe the steps in detail below.

\subsection{Candidate Rule Generation} \label{sec:rulemining}
The goal of this step is to generate promising candidate rules to create search space for the algorithm. A good rule should cover instances with a large treatment effect.
Our rule generation approach modifies the Virtual Twins (VT) methods \citep{foster_subgroup_2011}.
First, two random forests are built to predict $p(y_i=1)$ in control and treatment groups, respectively. 
\begin{equation}\label{eqn:pi}
\hat{p}_i(T_i) = p(y_i=1|T_i,\mathbf{x}_i) \text{ for } T_i = 0,1. \\
\end{equation}
Therefore, $\hat{p}_i(1) - \hat{p}_i(0)$ can be regarded as an estimate of the treatment effect on  instance $i$. We do not need this estimate to be accurate, as it is only used to provide potentially useful starting points for our method.
We proceed to define a binary variable to indicate if $\hat{p}_i(1) - \hat{p}_i(0)$ is larger than $\epsilon$:
\begin{equation}
Z_i = \mathbbm{1}\big(\hat{p}_i(1) - \hat{p}_i(0)> \epsilon\big),
\end{equation}
where $\epsilon$ is the average treatment effect on the entire dataset estimated using the implementation from \cite{sekhon2011multivariate}.
Thus we obtain a list of binary labels indicating whether  an instance is estimated to demonstrate treatment effect higher than the whole population. We then use the FP-growth algorithm \citep{han2000mining} to mine rules from the subset of training data with $Z_i=1$.  \citep[Note that there are many other off-the-shelf techniques that can be used instead of FP-growth if desired, see][]{michalski2013machine}.  In this particular work, we used the FP-Growth implementation in python of \citet{borgelt2005implementation}.
It takes binary-coded data where each column represents whether the attributes satisfy a condition. A condition can refer to either a continuous attribute within a range (for example, age is between 10 to 20) or a categorical attribute equal to a specific category (for example, gender is male). 

Since VT was developed to work on randomized data, whereas we assume we are working on observational data, here we do a balance check for each candidate rule to ensure the independence between $T$ and $\mathbf{x}$. We apply a multivariate two-sample Student's t-test on the treatment and control groups covered by each rule. It checks if two distributions have different means. If the means are statistically significantly different between treatment and control covariates, then this is not desired. We keep only rules with sufficiently large p-values. In our work we used $p>0.05$.

In practice, if too many rules remain, one can use the average treatment effect within each rule as a secondary criterion to further reduce the search space. This effectively induces the estimated average treatment effect to be a secondary desired goal for our desired rule sets. In our work, we evaluate the average treatment effect to further screen rules and keep the top $N$ rules with the largest average treatment effect. In our experiments, we empirically set $N$ to 5000.

These rules become the rule space $\mathcal{A}$. Restricting the algorithm to search only within $\mathcal{A}$ greatly reduces the search space. Let us discuss stochastic optimization in this space next.

\subsection{Proposal Step}
Our stochastic optimization approach is a variation of stochastic coordinate descent and simulated annealing. 
Instead of always moving to the best neighbor as in coordinate descent, we choose between exploration and exploitation, proposing a random neighbor occasionally to avoid getting stuck at a local optimum.
Identifying good neighbors to move to is critical in our search algorithm, since this directly determines how quickly the algorithm finds the optimal solution. Our proposing function consists of two steps: \emph{proposing an action} and \emph{selecting a rule to perform the action on}. Ideally, we would evaluate the objective function $F(\cdot)$ on every neighbor and then propose the best one (fastest descent). But this is impractical in real applications, due to the high computational cost incurred by fitting a Bayesian logistic regression in evaluating $F(\cdot)$ for all neighbors. Therefore, we propose a substitute metric $Q(\cdot)$ that is computationally convenient while providing a reliable evaluation of each neighbor. We provide the intuition for designing the substitute metric in what follows.

\textbf{Substitute Metric.}
A key observation in our model is that a good rule set always tries to cover instances that could potentially obey $y_i(1) - y_i(0) = 1$. (All other instances have a non-positive treatment effect.) We assume consistency in the sense of $y_i = y_i(T_i)$ given that $T_i$ was assigned.  While we are unable to observe both potential outcomes to find out exactly which examples to include, we are able to determine, by observing one of the potential outcomes and the treatment assignment, which instances 
would preferably be excluded from the rule set we are constructing. 

To do that, we divide the data space into three regions based on $y_i$ and $T_i$:    
\begin{align}
\mathcal{E}_0 &= \{\mathbf{x}_i|T_i =0, y_i= 1\} \;\textrm{(positive outcome without treatment)}\\
\mathcal{E}_1 &= \{\mathbf{x}_i|T_i =1, y_i= 0\}\; \textrm{(negative outcome with treatment)}\\
\mathcal{U} &= \{\mathbf{x}_i|T_i = y_i\}\; \textrm{(unknown treatment effect)}. \label{eqn:u}
\end{align}
If $\mathbf{x}_i\in\mathcal{E}_0$ or $\mathcal{E}_1$, the treatment effect must be non-positive, since instances in  $\mathcal{E}_0$ already have good outcomes in the control group  ($y_i(0) = 1$) so treatment is \emph{not necessary} and instances in  $\mathcal{E}_1$ have bad outcomes under treatment ($y_i(1) = 0$) so the treatment is \emph{not useful}. 
On the other hand, if $\mathbf{x}_i\in \mathcal{U}$, it means the instance has a good outcome with treatment ($y_i(1) = 1$) or bad outcome without treatment ($y_i(0) = 0$). However, it is \emph{unknown} whether the treatment is effective without knowing the other potential outcome, which needs to be inferred further by the model.  

Therefore, a good subgroup should try to exclude as many data points from $\mathcal{E}_0\cup\mathcal{E}_1$ as possible. Following this intuition, we define a metric $Q(\cdot)$ to represent the percentage of instances covered by a rule set $A$ from $\mathcal{E}_0\cup\mathcal{E}_1$.
\begin{equation}\label{eqn:precision}Q(A) = \frac{|I_{A}\cap(\mathcal{E}_0\cup\mathcal{E}_1)|}{|I_A|},\end{equation}
where $ I_A: = \{i = 1, \cdots, n: A(\mathbf{x}_i) = 1\}$ represents the indices of observations in subgroup $A$. We would like $Q(A)$ to be small for our choice of $A$.

\textbf{Choosing an Action (``ADD,'' ``CUT,''  or ``REPLACE'').} When proposing the next state, 
neighbors can be partitioned into three sets, each generated by adding, cutting, or replacing a rule from the current rule set. Choosing a good action will narrow down the search within only one subset of neighbors. 
To do that, we 
define
\begin{align}\epsilon^{[t]}&= \{i|i\in \mathcal{E}_0\cup \mathcal{E}_1, A^{[t]}(\mathbf{x}_i) = 1\} \label{eqn:et} \;\;\;\textrm{(points desired to exclude).}
\end{align}
The rule set $A^{[t]}$ is updated iteratively.
At iteration $t$, an example $\mathbf{x}_k$ is drawn uniformly from $\epsilon^{[t]}\cup \mathcal{U}$. Let $\mathcal{R}_1(\mathbf{x}_k)$ represent a set of rules that $\mathbf{x}_k$ satisfies and $\mathcal{R}_0(\mathbf{x}_k)$ represent a set of rules that $\mathbf{x}_k$ does not satisfy. If $\mathbf{x}_k \in \epsilon^{[t]}$, it means $A^{[t]}$ covers wrong data and we then find a neighboring rule set that covers less, by \emph{cutting} or \emph{replacing} a rule from $A^{[t]} \cap \mathcal{R}_1(\mathbf{x}_k)$, with equal probability -- this is our action in that case. If $\mathbf{x}_k \in \mathcal{U}$, then as explained previously, we are not sure if  $\mathbf{x}_k$ should or should not be covered and it is worth exploring. Therefore, as our action, a new rule set is proposed by either adding a rule from $\mathcal{R}_1(\mathbf{x}_k)$, cutting a rule in $A^{[t]}$, or replacing a rule from $A^{[t]}$, where the action is chosen uniformly at random. Once the action (cut, add, or replace) is determined, we then choose a rule to perform the action.

\textbf{Choosing a Rule.} After determining our action, we evaluate $Q(\cdot)$ on all possible neighbors produced by performing the selected action. Then a choice is made between exploration, choosing a random rule,  and exploitation, choosing the best rule (with the smallest $Q(\cdot)$. We denote the probability of exploration as $q$. This randomness helps avoid local optima and helps the Markov Chain to converge to a global optimum. 

We detail the three actions below. 
\begin{itemize}
\item ADD: With probability $q$, draw $z^{[t]}$ randomly from $\mathcal{R}_1(\mathbf{x}_k)\backslash A^{[t]}$; with probability $1-q$, $z^{[t]} \in \underset{a\in\mathcal{R}_1(\mathbf{x}_k)\backslash A^{[t]}}{\arg\min}\; Q(A^{[t]} \cup a).$
 Then $A^{[t+1]} \leftarrow A^{[t]} \cup z^{[t]}$.
\item CUT: if $\mathbf{x}_k \in \epsilon^{[t]}$, employ CUT-1; otherwise, if $\mathbf{x}_k \in \mathcal{U}$, employ CUT-2. 
\begin{itemize}
    \item CUT-1: with probability $q$, draw $z^{[t]}$ randomly from $A^{[t]}\cap \mathcal{R}_1(\mathbf{x}_k)$; with probability $1-q$, $z^{[t]} \in \underset{a\in A^{[t]}\cap \mathcal{R}_1(\mathbf{x}_k)}{\arg\min} Q(A^{[t]} \backslash a)$.  Then $A^{[t+1]} \leftarrow A^{[t]} \backslash z^{[t]}$. 
    \item  CUT-2: with probability $q$, draw $z^{[t]}$ randomly from $A^{[t]}$; with probability $1-q$, $z^{[t]} \in \underset{a\in A^{[t]}}{\arg\min} Q(A^{[t]} \backslash a)$.  Then $A^{[t+1]} \leftarrow A^{[t]} \backslash z^{[t]}$. 
\end{itemize} 
\item REPLACE: CUT, then ADD.
\end{itemize}
In summary, the proposal strategy assesses the current model and evaluates neighbors to ensure that, with a certain probability, the selected action and rule improve the current model. This is significantly more practically efficient than proposing moves uniformly at random.

\subsection{Conditional Optimality of a Hypothetical Rule Set}
 Note that $Q(\cdot)$ is a substitute metric that does not directly yield the optimal solution, i.e., a rule set with $Q(\cdot) = 0$ is not necessarily the optimal solution for the objective $F(\cdot)$. This is because $Q(\cdot)$ only encourages excluding data points from $\mathcal{E}_0\cup\mathcal{E}_1$ but cannot determine whether points from $\mathcal{U}$ should be included in a subgroup. $\mathcal{U}$ may also contain data points on which the treatment is not necessary ($y_i = 1, T_i = 1, y_i(0) = 1$) or not useful  ($y_i = 0, T_i = 0, y_i(1) = 0$). Here, we  investigate which conditions the model needs to satisfy such that $Q(\cdot)$ directly aligns with the objective of finding the optimal causal rule set. We first define a  
 \emph{Hypothetical Rule Set}.
\begin{definition}
Given a data set $\mathcal{D} = \{(\mathbf{x}_i,y_i,T_i)\}_{i=1}^n$, a \emph{Hypothetical Rule Set} $\bar{A}$ is defined as: 
$$\bar{A}(\mathbf{x}_i) = \mathbbm{1}(\mathbf{x}_i \in \mathcal{U}).$$
\end{definition}
A hypothetical rule set does not cover any point in  $\mathcal{E}_0\cup\mathcal{E}_1$. 

We prove in Theorem \ref{lm:CRS_Lup} that $\Theta(\mathcal{D};\bar{A},\mathbf{w}_{\bar{A}})$ is an upper bound on $\Theta(\cdot)$ (defined in Equation (\ref{eqn:theta})) for any rule set $A$  and its corresponding parameters $\mathbf{w}_A$, under the two constraints: $\gamma^{(1)}_A\leq0$ and $\gamma^{(1)}_A+\gamma^{(2)}_A \geq 0$.
\begin{theorem}\label{lm:CRS_Lup}
For any $A$, let $\mathbf{w}_A = \max_{\mathbf{w}}\Theta(A,\mathbf{w})$ and notate the elements in $\mathbf{w}_A$ as $\mathbf{w}_A = \{\mathbf{v}_A,\gamma^{(0)}_A,\gamma^{(1)}_A,\gamma^{(2)}_A\}$. If $\gamma^{(1)}_A\leq0, \gamma^{(1)}_A+\gamma^{(2)}_A \geq 0$, $$\Theta(A,\mathbf{w}_A) \leq \Theta(\bar{A},\mathbf{w}_{\bar{A}}).$$
\end{theorem}
See the supplementary material for the proof.
This theorem states that a Hypothetical Rule Set is an optimal solution to $F(A,\mathbf{w})$ under two constraints (ignoring the prior probability $\text{Prior}(A)$, which is not included in $\Theta$).  Note that this theorem only provides a theoretical upper bound on $\Theta(\cdot)$. It does not guarantee that a Hypothetical Rule Set is a feasible model since there might not exist a rule set that only covers examples in $\mathcal{U}$.
However, Theorem \ref{lm:CRS_Lup} illuminates criteria for evaluating a rule set, that a good rule set needs to cover much of $\mathcal{U}$ and little of $\mathcal{E}_0\cup\mathcal{E}_1$. 

\textbf{Remark}: Note that the above theorem holds only when  $\gamma^{(1)}_A+\gamma^{(2)}_A \geq 0$ and $\gamma^{(1)}_A\leq0$.
$\gamma^{(1)}_A+\gamma^{(2)}_A \geq 0$ ensures that the treatment effect for   the subgroup captured by $A$ should be positive. $\gamma^{(1)}_A\leq0$ means that being in subgroup $A$ alone does not lead to an elevated risk of $y_i(0) = 1$. Intuitively, keeping the baseline risk for subgroup $A$ low is helpful to avoid saturating the risks; when the input to function $\sigma$ is large from the baseline risk of $A$, even if the contribution $\gamma^{(2)}_A$ to the treatment effect from $A$ is large, little boost in treatment effect might be observed after the $\sigma$ transformation. Regardless of the practical intuition for this constraint, the theorem provides intuition for when $Q$ aligns with the goals of our true objective: namely, to cover points in $\mathcal{U}$ and not $\mathcal{E}_0\cup\mathcal{E}_1$. 


The conditional optimality implies that $Q$ is a reasonable metric but cannot be used to directly evaluate a rule set model to determine whether to accept or reject the proposal. Thus we only use $Q$ to evaluate rules when performing an action (CUT, ADD, or REPLACE). To avoid being entirely guided by $Q$, we add randomness in the rule selection step, and with probability $q$, a rule is randomly selected, as described above. The proposed solution is still evaluated by $F(\cdot)$ to determine whether to accept it.

\subsection{Constraint-Region Search}
We wish to further reduce the search complexity by utilizing properties derived from the model. 
In our proposed framework, the Bayesian prior places a preference on particular sizes of models (smaller models are preferred over large models for interpretability purposes), which means the MAP solution is likely to have few rules. Therefore,  we wish to derive an upper bound on the size of a MAP model, which can make the search algorithm more efficient since the space of models with few rules is smaller than the space of all rule sets. 
Furthermore, as we obtain better solutions throughout the search, we would like the bound to become tighter so that the search space becomes smaller as we narrow in on a MAP model.


Let $v^{[t]}$ denote the maximum objective value that we have seen at or before iteration $t$, $$v^{[t]} = \max_{\tau\leq t}F(A^{[\tau]}, \mathbf{w}_{A^{[\tau]}}).$$ Let $M^*_l$ represent the number of rules of length $l$ in the MAP model $A^*$ and let $m_l^{[t]}$ represent the derived upper bound on $M^*_l$ at iteration $t$. $m^{[t]}_l$ decreases monotonically with $t$ and is updated via Theorem \ref{thm:iterative} below.
\begin{theorem}\label{thm:iterative}
On dataset $\mathcal{D}$, apply a Causal Rule Set model with parameters $$H = \{\mathcal{A},L,\{\alpha_l,\beta_l\}_{l=1}^L,\Sigma, \mu\},$$ where $L,\{\alpha_l,\beta_l\}_{l=1,...L} \in \mathbb{N}^+$. Define $\{A^*,\mathbf{w}^*\} \in \arg\max_{A,\mathbf{w}} F(A,\mathbf{w})$ and $\mathbf{w}_{A^*} = \{\mathbf{v}_{A^*}$, $\gamma^{(0)}_{A^*}$, $\gamma^{(1)}_{A^*},\gamma^{(2)}_{A^*}\} $. If $\alpha_l<\beta_l, \gamma^{(1)}_{A^*}\leq 0,\gamma^{(1)}_{A^*}+\gamma^{(2)}_{A^*}\geq 0$, we have:  
\begin{equation*}
|A^*|\leq \sum_{l=1}^L   m_l^{[t]},\text{ where}
\end{equation*}
$$m_l^{[t]}=\frac{\log\Theta(\bar{A},\mathbf{w}_{\bar{A}})+\log \text{Prior}(\emptyset)-v^{[t]} }{\log \left(\frac{|\mathcal{A}_l|+\beta_{l}-1}{m_l^{[t-1]}+\alpha_{l}-1}\right)}$$ for $t\in \mathbbm{N}^+$ and $m_l^{[0]} = \frac{\log\Theta(\bar{A},\mathbf{w}_{\bar{A}})- \log\Theta(\emptyset,\mathbf{w}_{\emptyset})}{\log\left( \frac{|\mathcal{A}_l|+\beta_{l}-1}{|\mathcal{A}_l|+\alpha_{l}-1}\right)}.$
\end{theorem}

For Theorem \ref{thm:iterative},
the smaller $\frac{\alpha_l}{\beta_l}$, the tighter  the bound, which is consistent with the intuition of selecting rules of length $l$ with smaller $p_l$.  At time 0, we use an empty rule set as a benchmark, i.e., $v^{[0]} = \log\Theta(\emptyset,\mathbf{w}_\emptyset) + \log \text{Prior}(\emptyset),
m_l^{[-1]} \leq |\mathcal{A}_l|$,
yielding $m_l^{[0]}$ as above.
If an empty set is a good approximation, i.e., $ \Theta(\emptyset,\mathbf{w}_{\emptyset})$ is close to $\Theta(\bar{A},\mathbf{w}_{\bar{A}})$, then $|A^*|$ is small. This bound agrees with the intuition; if an empty set already achieves good performance, then adding more rules will likely hurt the outcome.

\begin{figure}[ht]\vspace{-2mm}
\centering
\includegraphics[width=0.65\textwidth]{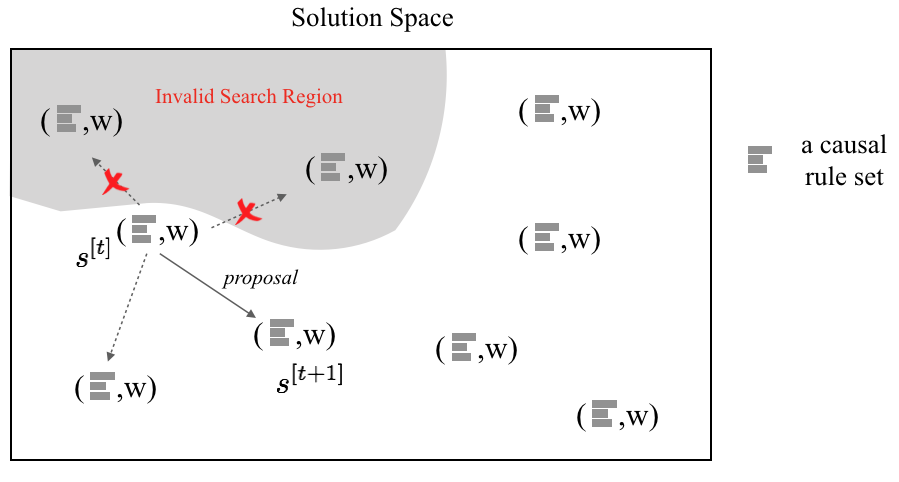}
\caption{Actively pruning the search space during the search. Each solution consists of an intermediate causal rule set and the corresponding parameter $w$. When proposing the next state, some region in the search space is pruned to improve the search efficiency}\label{fig:searchspace}
\end{figure}

As the search continues, $v^{[t]}$ becomes larger, and the upper bound $m_l^{[t]}$ decreases accordingly, pointing to a refined area of smaller sets of rules where our search algorithm should further explore. Therefore, in our algorithm, the probability of adding rules decays as $m_l^{[t]}$ decreases. The smaller this bound, the more often a CUT action should be selected instead of an ADD or REPLACE, in order to reduce the model size.  This bound lures the Markov chain towards regions of smaller models and helps to find the MAP model much more quickly. See Figure \ref{fig:searchspace} for an illustration of space pruning in proposing the next state.

Note that like Theorem 1, Theorem 2 holds under the two constraints that $\gamma_{A^*}^{(1)}\leq 0$ and $\gamma_{A^*}^{(1)}+\gamma_{A^*}^{(2)}\geq 0$, which may not hold in practice. However, when they do, the search becomes much more efficient. When the constraints are violated, the strategies we define will push the algorithm to select the action ADD less often, thus it may converge more slowly. One can adjust the impact of the moderating term $\exp(-\frac{\sum_{l=1}^L   m_l^{[t]}}{C})$ by adjusting $C$.

See the full algorithm in Algorithm \ref{alg:searchCRS}. We empirically set $C = 500$ and $q = 0.2$ in all experiments.
\begin{algorithm}[ht]
\caption{Search Algorithm for CRS}\label{alg:searchCRS}
\begin{algorithmic}[1]
\State Input: $C, q$
\State Initialize: $A^{[0]},v^{[0]} \leftarrow \emptyset, F(\emptyset,\mathbf{w}_{\emptyset})$
 \For{$t = 0,...,N_\text{iter}$}
        \State $\mathbf{x}_i \leftarrow \text{ an example randomly drawn from } \epsilon^{[t]} \cup \mathcal{U}$ ($\epsilon^{[t]}$ and $\mathcal{U}$ are from Formula (\ref{eqn:et}) and (\ref{eqn:u})).
\If{$\mathbf{x}_i \in \epsilon^{[t]}$ }
\State $A^{[t+1]} = \begin{cases}
        \text{CUT}(A^{[t]},q)\text{ with prob }\frac{1}{2} \\
        \text{REPLACE}(A^{[t]},q)\text{ with prob }\frac{1}{2} \\
        \end{cases}$
\Else
\State $A^{[t+1]} = \begin{cases}
        \text{CUT}(A^{[t]},q)\text{ w/ prob }\frac{2}{3}-\frac{1}{3}\exp(-\frac{\sum_{l=1}^L   m_l^{[t]}}{C})  \\
        \text{ADD}(A^{[t]},q)\text{ w/ prob }\frac{1}{3}\exp(-\frac{\sum_{l=1}^L   m_l^{[t]}}{C}) \\
         \text{REPLACE}(A^{[t]},q)\text{ with prob }\frac{1}{3} \\
        \end{cases}$
                
\EndIf
\State $\mathbf{w}_{A^{[t+1]}} = \arg\max_\mathbf{w} \Theta(A^{[t+1]},\mathbf{w})$ $\longrightarrow$ compute the best parameters for rule set $A^{[t+1]}$
\State $(A^{[t+1]}, \mathbf{w}_{A^{[t+1]}}) = (A_{t},\mathbf{w}_{A^{[t]}})$ with probability $\max\left\{1,\exp\left(\frac{F(A^{[t+1]},\mathbf{w}_{A^{[t+1]}}) - F(A^{[t]},\mathbf{w}_{A^{[t]}})}{B(t)}\right)\right\}$

\EndFor
\State return $A^{[N_\text{iter}+1]},\mathbf{w}_{A^{[N_\text{iter}+1]}}$
\end{algorithmic}
\end{algorithm}

\section{Simulation Analysis}\label{sec:exp}
In this section, we discuss the detailed experimental evaluation of CRS using synthetic data where the ground truth is known. We evaluate the performance of CRS in recovering the correct subgroups in different settings in terms of treatment generation mechanism, true subgroups, and various noise levels. We also conduct an ablation study to validate the effectiveness of the strategies we create in the algorithm. 
\subsection{Recovery Performance on Synthetic Data}  To test how accurately and how quickly CRS identifies the subgroups, we apply CRS to simulated datasets where true subgroups with enhanced treatment effects are predefined. We then compare true subgroups with subgroups recovered by CRS.    

We describe the basic procedure of data generation and will later vary its components in Section 5.2 to generate other settings.
In the basic procedure, features $\{\mathbf{x}_i\}_{i=1}^n$ are randomly drawn from normal distributions and  binary treatments $\{T_i\}_{i=1}^n$ are randomly assigned with probability 0.5 to be 1. Next, $y_i(0)$ are generated from  a logistic regression function with parameters $\mathbf{v}$ randomly drawn from normal distributions. $y_i(0) = 1$ if $\sigma(\mathbf{v}\mathbf{x}_{i})\geq 0.5$ and $y_i(0) = 0$ otherwise.
Then we mine rules from $\{\mathbf{x}_i\}_{i=1}^n$ and randomly choose five rules to form a true rule set $A^*$.  To set the treatment to be effective only on the subgroup covered by $A^*$ and non-effective everywhere else, we set:
\begin{equation}
   y_i(1)=  \begin{cases} 1 & \textrm{ when } A^*(\mathbf{x}_i)=1 \\  0 & \textrm{ when } A^*(\mathbf{x}_i)=0.\end{cases}
\end{equation}  
Then for each instance $\mathbf{x}_i$, the observed label is generated by   
\begin{equation}
    y_i= (1-T_i)y_i(0) + T_iy_i(1).
\end{equation}
Using this setup, we created a simulated dataset $\{(\mathbf{x}_i,y_i,T_i)\}^n_i$ with $n$ instances and $J$ features, $n \in \{10k, 100k\}$ and $J \in \{50, 100\}$. For each simulated dataset, 30\% was held out for testing. On the remaining 70\%, we first mined rules with minimum support of 5\% (we are only interested in subgroups with non-negligible sizes) and $L = 3$. We then selected $m$ candidate rules as $\mathcal{A}$. We set the expected means $\mu$ to 0 and variance $\Sigma$ to an identity matrix. Since we favor small models, we set $\alpha_l = 1$  and $\beta_l = |\mathcal{A}_l|$ for all $l$. Finally, we ran the search algorithm for 150 steps. 

 We evaluated the discrepancy between the recovered subgroup and the true subgroup $A^*$ over runs of the algorithm. To quantify the discrepancy and its convergence, we evaluated the error rate of the best solution found till iteration $t$, $\hat{A}^{[t]}$ on the test set, i.e.,
\begin{equation}
    \text{Error rate}(\hat{A}^{[t]} ) = \frac{\sum_i^n[\hat{A}^{[t]}(\mathbf{x}_i)\neq A^*(\mathbf{x}_i)]}{n}.
\end{equation}

\begin{figure}[ht]
\centering
        \includegraphics[width = 0.65\textwidth]{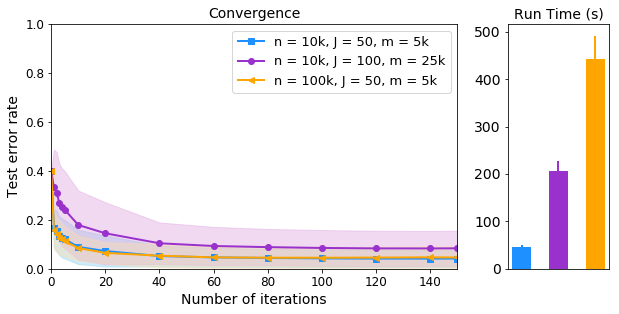}
        \caption{Convergence and runtime analysis of CRS.}\label{fig:convergence}
\end{figure}

We conducted three sets of experiments, varying $n, J$, and $m$. For each set of parameters, we repeated the experiment 100 times and recorded the error rate on the 30\% hold-out data each time. We plotted the average and standard deviation of the error rate in Figure \ref{fig:convergence}, together with their runtime in seconds.

Figure~\ref{fig:convergence} shows that \textbf{CRS recovers the true subgroup with high probability, and the convergence happens within 100 iterations}. Blue and yellow curves represent two sets of experiments on datasets with 50 features and different numbers of instances. We notice that these two curves almost entirely overlap. This is because the algorithm searches within the same size of rule space, 5000. 
On the other hand, if the search space increases, as it did for the third set of experiments where $m = 25000$, the algorithm had a slightly higher error rate since it is more difficult for the algorithm to search within a much larger space. 
While all three sets of experiments ran for 150 steps, their run time differed, which was determined by the number of candidate rules and the number of training instances.

\paragraph{Benchmarks} To compare with CRS, we also construct two benchmark algorithms for constructing a rule set. Both algorithms are greedy methods. They start with an empty rule set and iteratively include a new rule  based on some heuristics, evaluated on training data that are not captured by the current rule set. This type of greedy method has been popular in the past for building rule-based models for classification or regression purposes \citep{li2001cmar,yin2003cpar,chen2006new}, but often produce a large model with too many rules. We adopt a similar idea here and construct two heuristics. The first heuristic is the average treatment effect of instances covered by each rule, directly derived from its labels $Y_i$, which we call the ``greedy-hard'' method. The second heuristic uses the outputs from two random forests that estimate $p(y_i = 1)$ in control and treatment groups, respectively, i.e., Equation (\ref{eqn:pi}). Then we compute $p_i(1) - p_i(0)$ for each data point, and average them for each rule. Since the heuristic is based on the probabilistic output, we call this algorithm the ``greedy-soft'' method.

\begin{figure}[ht]
        \centering
        \includegraphics[width=\textwidth]{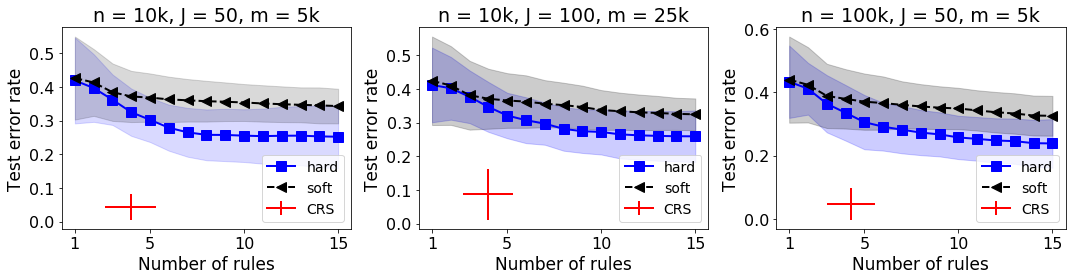}
        \caption{Out-of-sample performance of models of different sizes obtained by greedy benchmark algorithms.}\label{fig:benchmark}
\end{figure}

In both methods, we iteratively add the rules until reaching 15. Each experiment is repeated 100 times. We report in Figure \ref{fig:benchmark} the mean and standard deviation of the error evaluated on test sets.

Results show that the \textbf{performance of the benchmark algorithms is worse than CRS in all settings.} This is because the rule sets are formed via a greedy strategy without a global objective, while CRS considers the global fitness to the data as well as the complexity of the model (number of rules).
\subsection{Performance Analysis in Different Setups}
In this experiment, we varied different components in the simulation setup to examine the performance of CRS in various situations. In all experiments in this part, we set $n = 10000$, $J = 50$, and $m = 5000$. The data generation process is the same as in the basic procedure in Section 5.1, except components we modified otherwise. Each experiment was repeated 100 times in order to obtain the mean and standard deviations.

\begin{figure}[ht]
    \begin{subfigure}[t]{0.32\textwidth}
        \includegraphics[width = \textwidth]{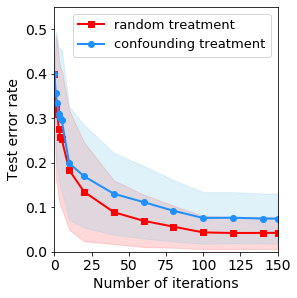}
        \caption{Performance under different treatment assignments.}
    \end{subfigure}%
    ~ 
    \begin{subfigure}[t]{0.32\textwidth}
        \centering
        \includegraphics[width=\textwidth]{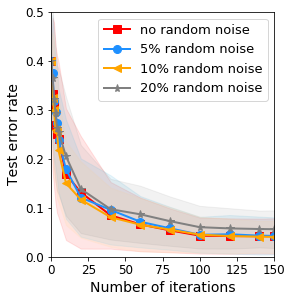}
        \caption{Performance under different noise levels.}
    \end{subfigure}
    ~
        \begin{subfigure}[t]{0.32\textwidth}
        \centering
        \includegraphics[width=\textwidth]{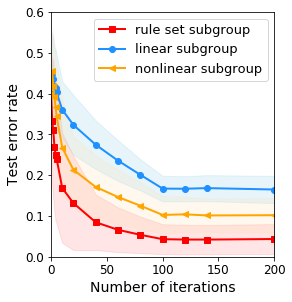}
        \caption{ Performance under different true subgroup generations.}
    \end{subfigure}
    \caption{Performance in different setups.}\label{fig:sim2}
\end{figure}

\paragraph{Varying the treatment generation.} In the previous simulations, the treatment was randomly assigned. We experimented with a more realistic setting where the treatment depends on the input.  To assign the treatment, a set of coefficients $\mathbf{w}_t$ were randomly drawn from normal distributions and $T_i = 1$ with probability
$\sigma(\mathbf{w}_t\mathbf{x}_i)$. The other parts of the experiment were the same as experiments in Section 5.1.
We then plotted the mean and standard deviation of the test error in Figure \ref{fig:sim2}(a). Results show that when the treatment is dependent on the input, CRS performs slightly worse than when the treatment is randomly assigned. 

\paragraph{Varying the noise level.}
To examine whether the method is robust to noise, we conducted a set of simulations where we randomly flipped 5\%, 10\% or 20\% of the true outcome labels in the training set, while keeping the other data generation the same as the basic procedure. We observe that the random noise does not have a significant impact on the performance. There was not a noticeable gap in the error until the noise was increased to 20\% of the data. This is because CRS relies on association rules to capture subgroups and each rule has a non-negligible support; thus these rules are fairly robust to random noise.

\paragraph{Varying the true subgroups.}
In the basic procedure, the true subgroups are covered by rule sets, which are consistent with the geometric coverage of CRS outputs. Here, we experiment with more ``shapes'' of subgroups, in particular, linear subgroups and certain types of nonlinear subgroups. To generate the linear subgroups, a set of coefficients $\mathbf{w}_\text{lin}$ is drawn from normal distributions and the true subgroup is defined as
\begin{equation}
    I_{A^*} = \{i|\sigma(\mathbf{w}_\text{lin}\mathbf{x}_i)\geq 0.5\}.
\end{equation}
Thus the subgroup covers a half-space of the feature space.
In addition, we generate nonlinear subgroups. To do that, we use feature engineering to add 10 polynomial terms of the original features, which are randomly set to be the square or cube of a randomly chosen feature, or the product of two randomly selected features. Let $\tilde{\mathbf{x}}_i$ represent the augmented feature vector. Then, we randomly generate a set of coefficients $\mathbf{w}_\text{nonlin}$ of the size $J+10$. The true subgroup is then defined as
\begin{equation}
    I_{A^*} = \{i|\sigma(\mathbf{w}_\text{nonlin}\tilde{\mathbf{x}}_i)\geq 0.5\}.
\end{equation}
Then we ran experiments following procedures described in Section 5.1.
The performance of CRS on different types of true subgroups is shown in Figure \ref{fig:sim2}(c). CRS performs better on non-linear subgroups than on linear subgroups. Note that the rule-set subgroups are also nonlinear. If the subgroups are defined by a linear boundary, then CRS can lose up to 20\% in predictive accuracy in our experiments. We could hypothesize that the worse performance on linear subgroups can be explained by the simple fact that an axis-aligned box is not a good model for a non-axis-aligned linear space.

\begin{figure*}[ht]
\centering
  \includegraphics[width=\textwidth]{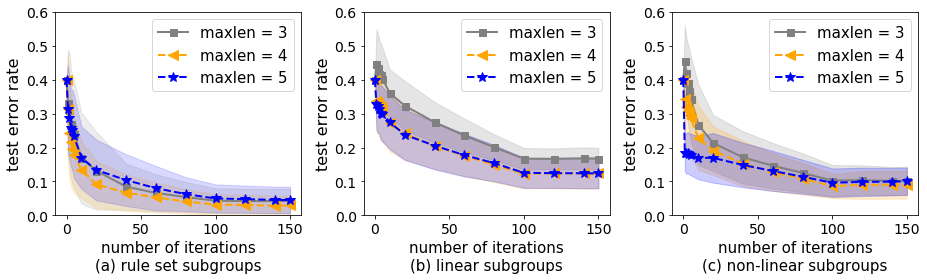}
\caption{Performance of CRS models with different maximum lengths of rules.}\label{fig:maxlen}
\end{figure*}

\paragraph{Varying the rule lengths.}  The length of rules influences the support and precision of rules. Rules with longer lengths will have smaller support and often higher precision. Longer rules are not only more complicated but also more easily to overfit. Therefore, in the previous analysis, we set the maximum length of rules to be 3, based on empirical experience from \cite{wang2017bayesian}. In this part, we increase the maximum length of rules to 4 and 5 and test CRS's performance under setups with different true subgroups described previously.  Figure \ref{fig:maxlen} shows that CRS 
models perform similarly with different maximum rule lengths if the ground truth subgroup is nonlinear (rule set subgroups are a special type of nonlinear subgroups). For the linear subgroups, longer rules achieve better performance: there is a nontrivial improvement when increasing the maximum length from 3 to 4 or 5.

\subsection{Ablation Study}
We conducted an ablation study for our search algorithm, where we removed some of the strategies we developed. We created three ablated versions of the original Algorithm \ref{alg:searchCRS}. 
\begin{enumerate}
    \item \textbf{CRS-ablated 1}: In Algorithm 1, an action was proposed based on a sample drawn from $\epsilon^{[t]} \cup \mathcal{U}$. We replace the action proposing strategy (lines 3 -- 8) in Algorithm 1 with random action selection: CUT, ADD, and REPLACE are each selected with  probability  $\frac{1}{3}$.
    \item \textbf{CRS-ablated 2}: In Algorithm 1, when performing an action, rules are evaluated with the substitution metric $Q(\cdot)$. We replace this strategy with random rule selection, i.e., increasing $q$ (the probability of selecting a random rule) to 1.
    \item \textbf{CRS-ablated 3}: In Algorithm 1, we generate candidate rules and screen the rules with a t-test  and average treatment effect.  We remove the rule-selection step so the CRS algorithm runs on a much larger set of candidate rules.
\end{enumerate}
\begin{figure*}[ht]
\centering
  \includegraphics[width=0.5\textwidth]{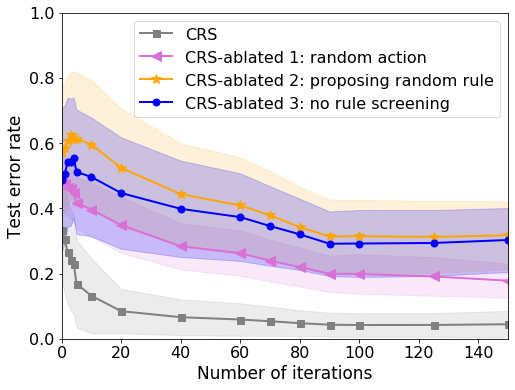}
\caption{Ablation study of the search algorithm for CRS}\label{fig:ablation}
\end{figure*}
Results show that the original algorithm performs significantly better than its ablated counterparts.

\section{Experiments on Real-world Datasets}
In this section, we evaluated the performance of CRS on real-world datasets from diverse domains where interpretability is highly valued by decision-makers, such as juvenile violence and consumer analysis.
\begin{table*}[ht]\renewcommand{\arraystretch}{1.3}
\centering
\caption{Summary of datasets}
\label{tab:data}
\small  
\begin{tabular}{c|c|c|l|c}
\toprule 
  &  \textbf{Treatment/Control}    & $\boldsymbol{n}$     & \multicolumn{1}{c}{\textbf{Descriptions }}    &$\boldsymbol{Y}$             \\ \hline
\multirow{4}{*}{\begin{tabular}[c]{@{}c@{}}In-vehicle\\ recommender\\ system\end{tabular}} & Direction& \multirow{4}{*}{12,684} & \multirow{4}{*}{\begin{tabular}[l]{@{}l@{}}User's attributes: gender, age,  marital\\ status, etc.; Contextual attributes:\\ destination, weather, time, etc.; Coupon's\\ attributes: time before it expires \end{tabular}} & \multirow{4}{*}{\begin{tabular}[l]{@{}l@{}}accept the\\ coupon \end{tabular}}\\
 &(same/opposite) & &   \\ \cline{2-2} 
&Price  & &   &  \\ 
 &($<$\$20/$\geq$\$20) &   &  \\ \hline
\multirow{3}{*}{Juvenile} & \multirow{3}{*}{\begin{tabular}[l]{@{}c@{}}Family member drinking \\  or drug problem (Y/N)\end{tabular}} & \multirow{3}{*}{4,023}     & \multirow{3}{*}{\begin{tabular}[l]{@{}l@{}}gender, age, friends delinquency, exposure  \\ to violence from friends or community, etc.\end{tabular}}    &   \multirow{3}{*}{delinquency}     \\                                                                                                                                                                                                             & & &  & \\ 
 & & &  & \\ \hline
\multirow{3}{*}{\begin{tabular}[c]{@{}c@{}}Young\\ Voter\\ Turnout\end{tabular}} & \multirow{3}{*}{\begin{tabular}[l]{@{}c@{}}Preregistration\\  (Y/N)\end{tabular}} & \multirow{3}{*}{5,562}     & \multirow{3}{*}{\begin{tabular}[l]{@{}l@{}}demographic features, party, median\\ household income, degree, district etc.\end{tabular}}    &   \multirow{3}{*}{turnout}     \\                                                                                                                                                                                                             & & &  & \\  
 & & &  & \\  \hline
\multirow{3}{*}{\begin{tabular}[l]{@{}c@{}}Medical \\crowdfunding\end{tabular}} & \multirow{3}{*}{\begin{tabular}[l]{@{}c@{}}First-person narration\\  (Y/N)\end{tabular}} & \multirow{3}{*}{51,228}     & \multirow{3}{*}{\begin{tabular}[l]{@{}l@{}}gender of patient, age of patient, length \\ of post, insurance, gender of fundraiser, \\  etc.\end{tabular}}    &   \multirow{3}{*}{\begin{tabular}[l]{@{}c@{}}crowdfunding \\ success\end{tabular} }     \\                                                                                                                                                                                                             & & &  & \\  
& & &  & \\  
 \bottomrule     
\end{tabular}
\end{table*}
\normalsize

   \subsection{Datasets Description}
The first data set is \emph{In-vehicle Recommender System} collected from Amazon Mechanical Turk via a survey \citep{wang2017bayesian}\footnote{The data are available at \url{https://github.com/wangtongada/BOA}}. 
 Turkers were asked whether they would accept a coupon under different driving scenarios characterized by the passenger, weather, destination, time, current location, etc. 
For this dataset, we conduct two experiments. In the first experiment, we choose the price range of average expenditure ($<$\$20 or $\geq$\$20) to be the treatment variable. The treatment $T=1$ if the price per person is less than \$20. In the second experiment, we choose the direction of the venue for using the coupon (whether it is in the same direction as the current destination) as the treatment and $T=1$ if it is in the same direction. In both experiments, the outcomes are whether drivers accept the recommended coupons. 

The second data set is \emph{Juvenile} dataset \citep{osofsky1995effect}, which was to study the effect of having family members with drinking or drug issues on juveniles committing delinquency. The data was collected via a survey sent to juveniles which asked several questions about whether they have witnessed violence in real life from their community, school, family, etc., whether their friends or family members have committed any delinquency (drug use, alcohol use, etc.), their demographic information, etc. The outcome was whether the respondents committed delinquency themselves. 

The third dataset is \emph{Youth Turnout} \citep{holbein2016making}, which was to examine the impact of preregistration on young voters (who have just turned 18) turnout for presidential election. The data was collected via a 2000-2012 Current Population Survey.

The last dataset is \emph{medical crowdfunding} data  provided by our industry partner. The data was collected from a medical crowdfunding platform and contains detailed case-level descriptions for over 51,000 crowdfunding cases. On this platform, each fundraiser submits information regarding a patient's age, gender, disease, etc., and then writes a short post calling for donations. We use this dataset to study if using first-person narration will improve the chance of success of the fundraising. Here, success is defined as receiving more than 30\% of the target amount.
See Table~\ref{tab:data} for a summary of the datasets. 

 We partitioned each data set into 60\% training, 20\% validation, and 20\% testing.

\subsection{Results}
\paragraph{Baselines.} We chose the following state-of-the-art baselines introduced in Section \ref{sec:related}: SIDES \citep{lipkovich_subgroup_2011}, VT \citep{foster_subgroup_2011} with a classification tree (VTC), VT with a regression tree (VTR), and QUINT \citep{dusseldorp2014qualitative}. The baseline models are tree-structured recursive partitioning methods. They differ in the heuristics the models use for greedy partitioning.  To generate treatment efficient frontiers for comparison with CRS, we set their maximum depths of trees to 3 to generate a list of models. We collected leaf nodes from the trees that correspond to estimated treatment effects larger than the average treatment effect. We then enumerated every possible combination of at most 3 leaves to create (a possibly large amount of) subgroups. Each selected subgroup corresponds to a rule or a rule combination (rule set). When running CRS, we vary parameters $\alpha_l, \beta_l$ to obtain models of various sizes

\paragraph{Obtaining the Treatment Efficient Frontier.}
Subgroups obtained by different methods are not directly comparable due to their differing sizes and treatment effects -- it cannot be determined if a smaller subgroup with a higher treatment effect is better or worse than a larger subgroup with a smaller treatment effect. Thus, we propose to generate a \emph{treatment efficient frontier} that characterizes the trade-off between the estimated treatment effect of the subgroup (y-axis) and the size of the subgroup (x-axis). This metric evaluates how quickly the estimated treatment effect magnitude will decay as the model recovers larger subgroups. We desire higher frontiers since they find better subgroups.

\begin{figure*}[ht]
\centering
  \includegraphics[width=0.95\textwidth]{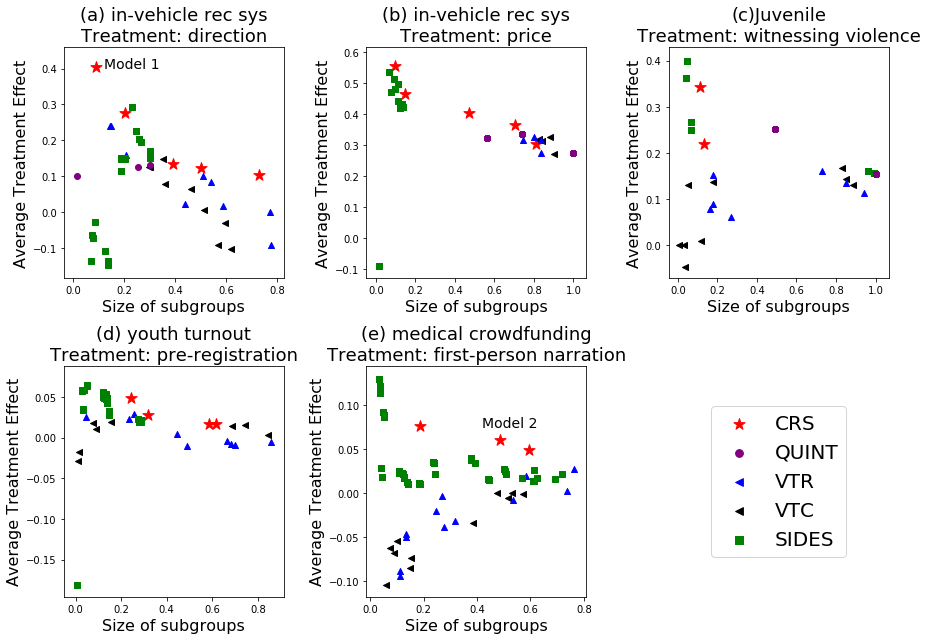}
\caption{\label{fig:experiments} Average treatment effect and the support of the subgroups discovered by CRS and baseline methods. CRS is designed to produce subgroups with elevated treatment effect, and a clear frontier is visible between treatment effect size and size of subgroups.}
\end{figure*}

 We evaluated the treatment effect and support of each of these subgroups on the validation set.  Subgroups on the Pareto-frontier were selected, such that none of the solutions were dominated by other solutions. Here ``dominate'' means a model is better than another model in both support and average treatment effect.  We then evaluated the size and average treatment effect selected models on the test set. See Figure~\ref{fig:experiments}.

In summary, \textbf{CRS achieved consistently competitive performance compared to baselines, as shown by higher treatment efficient frontiers in all experiments.} This means that compared to other baselines, CRS can accurately locate subgroups that can more strongly benefit from treatment. The result is unsurprising, given that it aims to globally optimize its objective, whereas baseline methods use greedy approaches.

\subsection{Examples}
For illustration and a more intuitive understanding of our model, we present two case studies from the above experiments.

\paragraph{Case Study on in-Vehicle Recommender System Data.}  We present an example of a CRS model (Model 1 in Figure~\ref{fig:experiments}), obtained from  the in-vehicle recommender system dataset with treatment ``direction.'' 
\begin{table}[ht]
\centering
\caption{A CRS model learned from the in-vehicle recommender system. The support is 8.9\% and the average treatment effect is 0.40, evaluated on the test set}\label{tab:coupon_CRS}
\begin{tabular}{llc}
\toprule
        & \multicolumn{1}{c}{\textbf{Rules}}         \\\hline
\textbf{If} &  weather $=$ snowy \textbf{\emph{and}} destination $\neq$ no immediate destination \textbf{\emph{and}} age $\leq$ 45   \\
& \textbf{\emph{OR}} weather $=$ snowy \textbf{\emph{and}} passenger $\neq$ friends  \\
\textbf{Then}        &  the instance is in the subgroup  \\ 
\textbf{Else}    &the instance is not in the subgroup      \\ \bottomrule                
\end{tabular}\vspace{-2mm}
\end{table}

This subgroup has the support of 8.9\% and an average treatment effect of 0.40.
It is interesting to notice that the two rules capture situations when people are reluctant to take a detour perhaps due to weather conditions (snowy) or/and when the driver is going to some immediate destination. In these cases, the relative location of a venue is of critical importance in creating the successful adoption of the coupon than other contexts. 

To compare with CRS, we also show the models obtained from other methods. For easy comparison, we show models closest to Model 1 in terms of support and treatment effect produced by each method. See Table \ref{tab:coupon_baseline}.
%
\begin{table}[ht]
\centering
\footnotesize
\caption{Rules or rule sets learned by baseline methods on in-vehicle recommender system dataset}\label{tab:coupon_baseline}
\begin{tabular}{l|lcc}
\toprule
\textbf{Methods}        &\multicolumn{1}{c}{\textbf{Rules}}  &support & ATE\\\hline
QUINT & passenger = kids \textbf{\emph{and}} destination = home & 0.01&0.1 \\ \hline
\multirow{2}{*}{VTC} &  destination $\neq $no immediate destination \textbf{\emph{and}} weather $\neq$ sunny  \textbf{\emph{and}} goes to a bar $\neq$ never &\multirow{2}{*}{0.29} & \multirow{2}{*}{0.13}  \\
& \textbf{\emph{OR}} destination $\neq $no immediate destination \textbf{\emph{and}} weather $=$ sunny \textbf{\emph{and}} destination $\neq$ work \\
\hline      
\multirow{3}{*}{VTR}&   destination $\neq $no immediate destination \textbf{\emph{and}} weather $=$ sunny  \textbf{\emph{and}} occupation $\neq$ sales related &\multirow{3}{*}{0.14} & \multirow{3}{*}{0.24}  \\
& \textbf{\emph{OR}} destination $\neq$ no immediate destination \textbf{\emph{and}} goes to restaurants with price per person &&\\
& of \$20 to \$50 $\neq$ never \textbf{\emph{and}} occupation = sales related & &\\ \hline
\multirow{3}{*}{SIDES}& passenger = None \textbf{\emph{and}} temperature = 30$\degree$F \textbf{\emph{and}} income $\geq$ \$25,000/year       &\multirow{3}{*}{ 0.23}&\multirow{3}{*}{0.29} \\
& \textbf{\emph{OR}} passenger = none \textbf{\emph{and}} time = 6pm \textbf{\emph{and}} age $\geq$ 21 &&\\
&  \textbf{\emph{OR}} destination = home \textbf{\emph{and}} temperature = 55$\degree$F \textbf{\emph{and}} time = 10 pm &&\\
\bottomrule                
\end{tabular}\vspace{-2mm}
\end{table}

Comparing the CRS model with baseline methods, we find that while the rules are not exactly the same, they turn out to use similar features, including weather, destination, and passenger.
\paragraph{Case Study on Medical Crowdfunding Data} We show another example of the CRS model obtained from the medical crowdfunding dataset (Model 2 in Figure \ref{fig:experiments}) in Table \ref{tab:crowdfunding} and a list of baseline models in Table \ref{tab:crowdfunding_baseline}
\begin{table}[ht]
\centering
\caption{A CRS model learned from the medical crowdfunding dataset. The support is 48.9\% and the average treatment effect is 0.06, evaluated on the test set}\label{tab:crowdfunding}
\begin{tabular}{llc}
\toprule
        & \multicolumn{1}{c}{\textbf{Rules}}         \\\hline
\textbf{If} &  patient age $\geq 44$ \textbf{\emph{and}} patient gender $\neq$ male \textbf{\emph{and}} the target amount $\leq$ 100k   \\
& \textbf{\emph{OR}} patient age $< 54$ \textbf{\emph{and}} month $=$ Jan or Feb \textbf{\emph{and}} writer of the post = patient  \\
& \textbf{\emph{OR}} status of the application = approved \textbf{\emph{and}} patient gender $\neq$ male \textbf{\emph{and}} target \\
&amount $<$ 300k  \\
\textbf{Then}        &  the instance is in the subgroup  \\ 
\textbf{Else}    &the instance is not in the subgroup      \\ \bottomrule                
\end{tabular}\vspace{-2mm}
\end{table}

\begin{table}[ht]
\centering
\footnotesize
\caption{Rules or rule sets learned by baseline methods for medical crowdfunding dataset. QUINT did not find any subgroups for this dataset}\label{tab:crowdfunding_baseline}
\begin{tabular}{c|lcc}
\toprule
\textbf{Methods}        & \multicolumn{1}{c}{\textbf{Rules}}  &support & ATE\\\hline
QUINT & None &-&- \\ \hline
\multirow{3}{*}{VTC} &  target amount $<$ 100k \textbf{\emph{and}} application $=$ approval  \textbf{\emph{and}} content length $<$ 213 &\multirow{3}{*}{0.48} & \multirow{3}{*}{0.00}  \\
& \textbf{\emph{OR}} target amount $\geq$ 100k \textbf{\emph{and}} patient age $<$ 44 \textbf{\emph{and}} content length $<$ 337 && \\
& \textbf{\emph{OR}} target amount $\geq$ 100k \textbf{\emph{and}} patient age $<$ 44  \textbf{\emph{and}} content length $\geq$ 337 &&\\
\hline      
\multirow{3}{*}{VTR}&   target amount $<$ 100k \textbf{\emph{and}} application = approved \textbf{\emph{and}} month = after March &\multirow{3}{*}{0.58} & \multirow{3}{*}{0.02}  \\
& \textbf{\emph{OR}} target amount $\geq$ 100k\textbf{\emph{and}} patient age $<$ 44 \textbf{\emph{and}} content length $\geq$ 522 &&\\
& \textbf{\emph{OR}}  target amount $\geq$ 100k \textbf{\emph{and}} patient age $\geq$ 44 \textbf{\emph{and}} month = after March  & &\\ \hline
\multirow{3}{*}{SIDES}& weekday = True \textbf{\emph{and}} application = approved \textbf{\emph{and}} has commercial insurance = False      &\multirow{3}{*}{ 0.50}&\multirow{3}{*}{0.03} \\
& \textbf{\emph{OR}} commercial insurance = True \textbf{\emph{and}} application = approved \textbf{\emph{and}} month = after March &&\\
&  \textbf{\emph{OR}} month = after June \textbf{\emph{and}} content length $\geq $ 213&&\\
\bottomrule                
\end{tabular}\vspace{-2mm}
\end{table}
Like the previous case study, the rules learned by the baselines share many common features with the rules in the CRS model, including the target amount, patient age, and month. See more examples included in the Appendix.

\section{Conclusion and Discussion}
We presented Causal Rule Sets for identifying a subgroup with an enhanced treatment effect, characterized by a small set of short rules.  Compared with previous recursive partitioning methods that use one rule to capture a group, CRS is able to capture subgroups of various shapes and sizes.  In addition, CRS does not use greedy partitioning but optimizes under a carefully designed global objective that balances the fitness to the data and model complexity through a generative framework. To the best of our knowledge, this is the first work to use a set of rules to capture a subgroup with an enhanced treatment effect using a principled non-greedy approach. The Bayesian logistic regression we proposed adopted a classical regression function for modeling potential outcomes,  capturing relationships between variables, treatment, and the subgroup. The inference method used theoretically grounded heuristics and bounding strategies to improve search efficiency. 

The machine learning model we develop extends the existing literature on causal subgroup discovery and has demonstrated better performance in terms of better treatment efficient frontiers. We envision our work will be useful for domains like personalized therapy and targeted advertising.

\subsection*{Future Work } There are several directions for future work. We have used a generative approach, but it is possible to derive a frequentist approach that leads to a similar optimization problem of finding an optimal subgroup with elevated treatment effect. Then, instead of making MAP inference, one can directly optimize an objective that considers the elevated treatment effect and the sparsity of the rule set (e.g., the number of rules in the model). Also, rather than finding a subgroup with an elevated treatment effect, one could directly model the treatment effect with a sparse piecewise constant treatment effect. This can be modeled as an optimal decision tree problem, for which modern approaches have made significant headway \citep{LinEtAl20,HuRuSe2019}, but have not yet been extended to causal analysis. (All of the baseline tree-based approaches we discussed have used greedy splitting and pruning for forming the trees, whereas the possible extension we mention would yield \emph{optimal} trees.) Our work could also be extended to handle decision analysis problems, where the subgroup identified would not only have elevated treatment effect but would also be cost-effective to treat, along the lines of \citet{lakkaraju2017learning} where a treatment regime could be learned. 
Finally, there are many possible direct applications of our work to real problems, such as determining which patients would respond to a vaccine or drug therapy, where interpretability of the model and identification of the group of patients with elevated treatment effects are paramount.

\bibliographystyle{ACM-Reference-Format}
\bibliography{causal} 


\begin{thebibliography}{59}


\ifx \showCODEN    \undefined \def \showCODEN     #1{\unskip}     \fi
\ifx \showDOI      \undefined \def \showDOI       #1{#1}\fi
\ifx \showISBNx    \undefined \def \showISBNx     #1{\unskip}     \fi
\ifx \showISBNxiii \undefined \def \showISBNxiii  #1{\unskip}     \fi
\ifx \showISSN     \undefined \def \showISSN      #1{\unskip}     \fi
\ifx \showLCCN     \undefined \def \showLCCN      #1{\unskip}     \fi
\ifx \shownote     \undefined \def \shownote      #1{#1}          \fi
\ifx \showarticletitle \undefined \def \showarticletitle #1{#1}   \fi
\ifx \showURL      \undefined \def \showURL       {\relax}        \fi
\providecommand\bibfield[2]{#2}
\providecommand\bibinfo[2]{#2}
\providecommand\natexlab[1]{#1}
\providecommand\showeprint[2][]{arXiv:#2}

\bibitem[\protect\citeauthoryear{Angelino, Larus-Stone, Alabi, Seltzer, and
  Rudin}{Angelino et~al\mbox{.}}{2017}]%
        {angelino2017learning}
\bibfield{author}{\bibinfo{person}{Elaine Angelino}, \bibinfo{person}{Nicholas
  Larus-Stone}, \bibinfo{person}{Daniel Alabi}, \bibinfo{person}{Margo
  Seltzer}, {and} \bibinfo{person}{Cynthia Rudin}.}
  \bibinfo{year}{2017}\natexlab{}.
\newblock \showarticletitle{Learning certifiably optimal rule lists for
  categorical data}.
\newblock \bibinfo{journal}{\emph{The Journal of Machine Learning Research}}
  \bibinfo{volume}{18}, \bibinfo{number}{1} (\bibinfo{year}{2017}),
  \bibinfo{pages}{8753--8830}.
\newblock


\bibitem[\protect\citeauthoryear{Angrist, Imbens, and Rubin}{Angrist
  et~al\mbox{.}}{1996}]%
        {angrist1996identification}
\bibfield{author}{\bibinfo{person}{Joshua~D Angrist}, \bibinfo{person}{Guido~W
  Imbens}, {and} \bibinfo{person}{Donald~B Rubin}.}
  \bibinfo{year}{1996}\natexlab{}.
\newblock \showarticletitle{Identification of causal effects using instrumental
  variables}.
\newblock \bibinfo{journal}{\emph{Journal of the American statistical
  Association}} \bibinfo{volume}{91}, \bibinfo{number}{434}
  (\bibinfo{year}{1996}), \bibinfo{pages}{444--455}.
\newblock


\bibitem[\protect\citeauthoryear{Assmann, Pocock, Enos, and Kasten}{Assmann
  et~al\mbox{.}}{2000}]%
        {assmann2000subgroup}
\bibfield{author}{\bibinfo{person}{Susan~F Assmann}, \bibinfo{person}{Stuart~J
  Pocock}, \bibinfo{person}{Laura~E Enos}, {and} \bibinfo{person}{Linda~E
  Kasten}.} \bibinfo{year}{2000}\natexlab{}.
\newblock \showarticletitle{Subgroup analysis and other (mis) uses of baseline
  data in clinical trials}.
\newblock \bibinfo{journal}{\emph{The Lancet}} \bibinfo{volume}{355},
  \bibinfo{number}{9209} (\bibinfo{year}{2000}), \bibinfo{pages}{1064--1069}.
\newblock


\bibitem[\protect\citeauthoryear{Atzmueller}{Atzmueller}{2015}]%
        {atzmueller2015subgroup}
\bibfield{author}{\bibinfo{person}{Martin Atzmueller}.}
  \bibinfo{year}{2015}\natexlab{}.
\newblock \showarticletitle{Subgroup discovery}.
\newblock \bibinfo{journal}{\emph{Wiley Interdisciplinary Reviews: Data Mining
  and Knowledge Discovery}} \bibinfo{volume}{5}, \bibinfo{number}{1}
  (\bibinfo{year}{2015}), \bibinfo{pages}{35--49}.
\newblock


\bibitem[\protect\citeauthoryear{Atzmueller and Puppe}{Atzmueller and
  Puppe}{2006}]%
        {atzmueller2006sd}
\bibfield{author}{\bibinfo{person}{Martin Atzmueller} {and}
  \bibinfo{person}{Frank Puppe}.} \bibinfo{year}{2006}\natexlab{}.
\newblock \showarticletitle{SD-Map--A fast algorithm for exhaustive subgroup
  discovery}. In \bibinfo{booktitle}{\emph{European Conference on Principles of
  Data Mining and Knowledge Discovery}}. Springer, \bibinfo{pages}{6--17}.
\newblock


\bibitem[\protect\citeauthoryear{Baselga, Perez, Pienkowski, and Bell}{Baselga
  et~al\mbox{.}}{2006}]%
        {baselga2006adjuvant}
\bibfield{author}{\bibinfo{person}{Jose Baselga}, \bibinfo{person}{Edith~A
  Perez}, \bibinfo{person}{Tadeusz Pienkowski}, {and} \bibinfo{person}{Richard
  Bell}.} \bibinfo{year}{2006}\natexlab{}.
\newblock \showarticletitle{Adjuvant trastuzumab: a milestone in the treatment
  of HER-2-positive early breast cancer}.
\newblock \bibinfo{journal}{\emph{The Oncologist}} \bibinfo{volume}{11},
  \bibinfo{number}{Supplement 1} (\bibinfo{year}{2006}),
  \bibinfo{pages}{4--12}.
\newblock


\bibitem[\protect\citeauthoryear{Borgelt}{Borgelt}{2005}]%
        {borgelt2005implementation}
\bibfield{author}{\bibinfo{person}{Christian Borgelt}.}
  \bibinfo{year}{2005}\natexlab{}.
\newblock \showarticletitle{An Implementation of the FP-growth Algorithm}. In
  \bibinfo{booktitle}{\emph{Proc of the 1st International Workshop on Open
  Source Data Mining: Frequent Pattern Mining Implementations}}. ACM,
  \bibinfo{pages}{1--5}.
\newblock


\bibitem[\protect\citeauthoryear{Brijain, Patel, Kushik, and Rana}{Brijain
  et~al\mbox{.}}{2014}]%
        {brijain2014survey}
\bibfield{author}{\bibinfo{person}{Mr Brijain}, \bibinfo{person}{R Patel},
  \bibinfo{person}{Mr Kushik}, {and} \bibinfo{person}{K Rana}.}
  \bibinfo{year}{2014}\natexlab{}.
\newblock \showarticletitle{A survey on decision tree algorithm for
  classification}.
\newblock \bibinfo{journal}{\emph{International Journal of Engineering
  Developmentand Research}} \bibinfo{volume}{2}, \bibinfo{number}{1}
  (\bibinfo{year}{2014}).
\newblock


\bibitem[\protect\citeauthoryear{Carmona, Gonz{\'a}lez, del Jesus, and
  Herrera}{Carmona et~al\mbox{.}}{2009}]%
        {carmona2009analysis}
\bibfield{author}{\bibinfo{person}{Crist{\'o}bal~Jos{\'e} Carmona},
  \bibinfo{person}{Pedro Gonz{\'a}lez}, \bibinfo{person}{Maria~Jose del Jesus},
  {and} \bibinfo{person}{Francisco Herrera}.} \bibinfo{year}{2009}\natexlab{}.
\newblock \showarticletitle{An analysis of evolutionary algorithms with
  different types of fuzzy rules in subgroup discovery}. In
  \bibinfo{booktitle}{\emph{Fuzzy Systems, 2009. FUZZ-IEEE 2009. IEEE
  International Conference on}}. IEEE, \bibinfo{pages}{1706--1711}.
\newblock


\bibitem[\protect\citeauthoryear{Chen, Liu, Yu, Wei, and Zhang}{Chen
  et~al\mbox{.}}{2006}]%
        {chen2006new}
\bibfield{author}{\bibinfo{person}{Guoqing Chen}, \bibinfo{person}{Hongyan
  Liu}, \bibinfo{person}{Lan Yu}, \bibinfo{person}{Qiang Wei}, {and}
  \bibinfo{person}{Xing Zhang}.} \bibinfo{year}{2006}\natexlab{}.
\newblock \showarticletitle{A new approach to classification based on
  association rule mining}.
\newblock \bibinfo{journal}{\emph{Decision Support Systems}}
  \bibinfo{volume}{42}, \bibinfo{number}{2} (\bibinfo{year}{2006}),
  \bibinfo{pages}{674--689}.
\newblock


\bibitem[\protect\citeauthoryear{Chetty, Hendren, and Katz}{Chetty
  et~al\mbox{.}}{2016}]%
        {chetty2016effects}
\bibfield{author}{\bibinfo{person}{Raj Chetty}, \bibinfo{person}{Nathaniel
  Hendren}, {and} \bibinfo{person}{Lawrence~F Katz}.}
  \bibinfo{year}{2016}\natexlab{}.
\newblock \showarticletitle{The effects of exposure to better neighborhoods on
  children: New evidence from the moving to opportunity experiment}.
\newblock \bibinfo{journal}{\emph{American Economic Review}}
  \bibinfo{volume}{106}, \bibinfo{number}{4} (\bibinfo{year}{2016}),
  \bibinfo{pages}{855--902}.
\newblock


\bibitem[\protect\citeauthoryear{Cleophas, Zwinderman, and Cleophas}{Cleophas
  et~al\mbox{.}}{2002}]%
        {cleophas2002subgroup}
\bibfield{author}{\bibinfo{person}{Ton~J Cleophas}, \bibinfo{person}{Aeilko~H
  Zwinderman}, {and} \bibinfo{person}{Toine~F Cleophas}.}
  \bibinfo{year}{2002}\natexlab{}.
\newblock \showarticletitle{Subgroup analysis using multiple linear regression:
  confounding, interaction, synergism}.
\newblock In \bibinfo{booktitle}{\emph{Statistics Applied to Clinical Trials}}.
  \bibinfo{publisher}{Springer}, \bibinfo{pages}{95--104}.
\newblock


\bibitem[\protect\citeauthoryear{Cohen, Cohen, West, and Aiken}{Cohen
  et~al\mbox{.}}{2013}]%
        {cohen2013applied}
\bibfield{author}{\bibinfo{person}{Jacob Cohen}, \bibinfo{person}{Patricia
  Cohen}, \bibinfo{person}{Stephen~G West}, {and} \bibinfo{person}{Leona~S
  Aiken}.} \bibinfo{year}{2013}\natexlab{}.
\newblock \bibinfo{booktitle}{\emph{Applied multiple regression/correlation
  analysis for the behavioral sciences}}.
\newblock \bibinfo{publisher}{Routledge}.
\newblock


\bibitem[\protect\citeauthoryear{Cook}{Cook}{1971}]%
        {cook1971complexity}
\bibfield{author}{\bibinfo{person}{Stephen~A Cook}.}
  \bibinfo{year}{1971}\natexlab{}.
\newblock \showarticletitle{The complexity of theorem-proving procedures}. In
  \bibinfo{booktitle}{\emph{Proceedings of the third annual ACM symposium on
  Theory of computing}}. ACM, \bibinfo{pages}{151--158}.
\newblock


\bibitem[\protect\citeauthoryear{Dash, Gunluk, and Wei}{Dash
  et~al\mbox{.}}{2018}]%
        {dash2018boolean}
\bibfield{author}{\bibinfo{person}{Sanjeeb Dash}, \bibinfo{person}{Oktay
  Gunluk}, {and} \bibinfo{person}{Dennis Wei}.}
  \bibinfo{year}{2018}\natexlab{}.
\newblock \showarticletitle{Boolean decision rules via column generation}. In
  \bibinfo{booktitle}{\emph{Advances in Neural Information Processing
  Systems}}. \bibinfo{pages}{4655--4665}.
\newblock


\bibitem[\protect\citeauthoryear{Del~Jesus, Gonz{\'a}lez, Herrera, and
  Mesonero}{Del~Jesus et~al\mbox{.}}{2007}]%
        {del2007evolutionary}
\bibfield{author}{\bibinfo{person}{Mar{\'\i}a~Jos{\'e} Del~Jesus},
  \bibinfo{person}{Pedro Gonz{\'a}lez}, \bibinfo{person}{Francisco Herrera},
  {and} \bibinfo{person}{Mikel Mesonero}.} \bibinfo{year}{2007}\natexlab{}.
\newblock \showarticletitle{Evolutionary fuzzy rule induction process for
  subgroup discovery: a case study in marketing}.
\newblock \bibinfo{journal}{\emph{IEEE Transactions on Fuzzy Systems}}
  \bibinfo{volume}{15}, \bibinfo{number}{4} (\bibinfo{year}{2007}),
  \bibinfo{pages}{578--592}.
\newblock


\bibitem[\protect\citeauthoryear{Dusseldorp and Van~Mechelen}{Dusseldorp and
  Van~Mechelen}{2014}]%
        {dusseldorp2014qualitative}
\bibfield{author}{\bibinfo{person}{Elise Dusseldorp} {and}
  \bibinfo{person}{Iven Van~Mechelen}.} \bibinfo{year}{2014}\natexlab{}.
\newblock \showarticletitle{Qualitative interaction trees: a tool to identify
  qualitative treatment--subgroup interactions}.
\newblock \bibinfo{journal}{\emph{Statistics in Medicine}}
  \bibinfo{volume}{33}, \bibinfo{number}{2} (\bibinfo{year}{2014}),
  \bibinfo{pages}{219--237}.
\newblock


\bibitem[\protect\citeauthoryear{Figlio, Guryan, Karbownik, and Roth}{Figlio
  et~al\mbox{.}}{2014}]%
        {figlio2014effects}
\bibfield{author}{\bibinfo{person}{David Figlio}, \bibinfo{person}{Jonathan
  Guryan}, \bibinfo{person}{Krzysztof Karbownik}, {and}
  \bibinfo{person}{Jeffrey Roth}.} \bibinfo{year}{2014}\natexlab{}.
\newblock \showarticletitle{The effects of poor neonatal health on children's
  cognitive development}.
\newblock \bibinfo{journal}{\emph{American Economic Review}}
  \bibinfo{volume}{104}, \bibinfo{number}{12} (\bibinfo{year}{2014}),
  \bibinfo{pages}{3921--55}.
\newblock


\bibitem[\protect\citeauthoryear{Foster, Taylor, and Ruberg}{Foster
  et~al\mbox{.}}{2011}]%
        {foster_subgroup_2011}
\bibfield{author}{\bibinfo{person}{Jared~C Foster}, \bibinfo{person}{Jeremy~MG
  Taylor}, {and} \bibinfo{person}{Stephen~J Ruberg}.}
  \bibinfo{year}{2011}\natexlab{}.
\newblock \showarticletitle{Subgroup identification from randomized clinical
  trial data}.
\newblock \bibinfo{journal}{\emph{Statistics in Medicine}}
  \bibinfo{volume}{30}, \bibinfo{number}{24} (\bibinfo{year}{2011}),
  \bibinfo{pages}{2867--2880}.
\newblock


\bibitem[\protect\citeauthoryear{Gamberger and Lavrac}{Gamberger and
  Lavrac}{2002}]%
        {gamberger2002expert}
\bibfield{author}{\bibinfo{person}{Dragan Gamberger} {and}
  \bibinfo{person}{Nada Lavrac}.} \bibinfo{year}{2002}\natexlab{}.
\newblock \showarticletitle{Expert-guided subgroup discovery: Methodology and
  application}.
\newblock \bibinfo{journal}{\emph{Journal of Artificial Intelligence Research}}
   \bibinfo{volume}{17} (\bibinfo{year}{2002}), \bibinfo{pages}{501--527}.
\newblock


\bibitem[\protect\citeauthoryear{Han, Pei, and Yin}{Han et~al\mbox{.}}{2000}]%
        {han2000mining}
\bibfield{author}{\bibinfo{person}{Jiawei Han}, \bibinfo{person}{Jian Pei},
  {and} \bibinfo{person}{Yiwen Yin}.} \bibinfo{year}{2000}\natexlab{}.
\newblock \showarticletitle{Mining frequent patterns without candidate
  generation}.
\newblock \bibinfo{journal}{\emph{ACM sigmod record}} \bibinfo{volume}{29},
  \bibinfo{number}{2} (\bibinfo{year}{2000}), \bibinfo{pages}{1--12}.
\newblock


\bibitem[\protect\citeauthoryear{Holbein and Hillygus}{Holbein and
  Hillygus}{2016}]%
        {holbein2016making}
\bibfield{author}{\bibinfo{person}{John~B Holbein} {and}
  \bibinfo{person}{D~Sunshine Hillygus}.} \bibinfo{year}{2016}\natexlab{}.
\newblock \showarticletitle{Making young voters: the impact of preregistration
  on youth turnout}.
\newblock \bibinfo{journal}{\emph{American Journal of Political Science}}
  \bibinfo{volume}{60}, \bibinfo{number}{2} (\bibinfo{year}{2016}),
  \bibinfo{pages}{364--382}.
\newblock


\bibitem[\protect\citeauthoryear{Hu, Rudin, and Seltzer}{Hu
  et~al\mbox{.}}{2019}]%
        {HuRuSe2019}
\bibfield{author}{\bibinfo{person}{Xiyang Hu}, \bibinfo{person}{Cynthia Rudin},
  {and} \bibinfo{person}{Margo Seltzer}.} \bibinfo{year}{2019}\natexlab{}.
\newblock \showarticletitle{Optimal sparse decision trees}. In
  \bibinfo{booktitle}{\emph{Advances in Neural Information Processing
  Systems}}. \bibinfo{pages}{7267--7275}.
\newblock


\bibitem[\protect\citeauthoryear{Kav{\v{s}}ek and Lavra{\v{c}}}{Kav{\v{s}}ek
  and Lavra{\v{c}}}{2006}]%
        {kavvsek2006apriori}
\bibfield{author}{\bibinfo{person}{Branko Kav{\v{s}}ek} {and}
  \bibinfo{person}{Nada Lavra{\v{c}}}.} \bibinfo{year}{2006}\natexlab{}.
\newblock \showarticletitle{APRIORI-SD: Adapting association rule learning to
  subgroup discovery}.
\newblock \bibinfo{journal}{\emph{Applied Artificial Intelligence}}
  \bibinfo{volume}{20}, \bibinfo{number}{7} (\bibinfo{year}{2006}),
  \bibinfo{pages}{543--583}.
\newblock


\bibitem[\protect\citeauthoryear{Kim}{Kim}{2016}]%
        {kim2016hybrid}
\bibfield{author}{\bibinfo{person}{Kyoungok Kim}.}
  \bibinfo{year}{2016}\natexlab{}.
\newblock \showarticletitle{A hybrid classification algorithm by subspace
  partitioning through semi-supervised decision tree}.
\newblock \bibinfo{journal}{\emph{Pattern Recognition}}  \bibinfo{volume}{60}
  (\bibinfo{year}{2016}), \bibinfo{pages}{157--163}.
\newblock


\bibitem[\protect\citeauthoryear{Lagakos}{Lagakos}{2006}]%
        {lagakos2006challenge}
\bibfield{author}{\bibinfo{person}{Stephen~W Lagakos}.}
  \bibinfo{year}{2006}\natexlab{}.
\newblock \showarticletitle{The challenge of subgroup analyses-reporting
  without distorting}.
\newblock \bibinfo{journal}{\emph{New England Journal of Medicine}}
  \bibinfo{volume}{354}, \bibinfo{number}{16} (\bibinfo{year}{2006}),
  \bibinfo{pages}{1667}.
\newblock


\bibitem[\protect\citeauthoryear{Lakkaraju, Bach, and Leskovec}{Lakkaraju
  et~al\mbox{.}}{2016}]%
        {lakkaraju2016interpretable}
\bibfield{author}{\bibinfo{person}{Himabindu Lakkaraju},
  \bibinfo{person}{Stephen~H Bach}, {and} \bibinfo{person}{Jure Leskovec}.}
  \bibinfo{year}{2016}\natexlab{}.
\newblock \showarticletitle{Interpretable decision sets: A joint framework for
  description and prediction}. In \bibinfo{booktitle}{\emph{ACM SIGKDD}}. ACM,
  \bibinfo{pages}{1675--1684}.
\newblock


\bibitem[\protect\citeauthoryear{Lakkaraju and Rudin}{Lakkaraju and
  Rudin}{2017}]%
        {lakkaraju2017learning}
\bibfield{author}{\bibinfo{person}{Himabindu Lakkaraju} {and}
  \bibinfo{person}{Cynthia Rudin}.} \bibinfo{year}{2017}\natexlab{}.
\newblock \showarticletitle{Learning cost-effective and interpretable treatment
  regimes}. In \bibinfo{booktitle}{\emph{Artificial Intelligence and
  Statistics}}. \bibinfo{pages}{166--175}.
\newblock


\bibitem[\protect\citeauthoryear{Lavrac, Kavsek, Flach, and Todorovski}{Lavrac
  et~al\mbox{.}}{2004}]%
        {lavrac2004subgroup}
\bibfield{author}{\bibinfo{person}{Nada Lavrac}, \bibinfo{person}{Branko
  Kavsek}, \bibinfo{person}{Peter Flach}, {and} \bibinfo{person}{Ljupco
  Todorovski}.} \bibinfo{year}{2004}\natexlab{}.
\newblock \showarticletitle{Subgroup discovery with CN2-SD}.
\newblock \bibinfo{journal}{\emph{J. Mach. Learn. Res.}} \bibinfo{volume}{5},
  \bibinfo{number}{2} (\bibinfo{year}{2004}), \bibinfo{pages}{153--188}.
\newblock


\bibitem[\protect\citeauthoryear{Lee, Bargagli-Stoffi, and Dominici}{Lee
  et~al\mbox{.}}{2020}]%
        {lee2020causal}
\bibfield{author}{\bibinfo{person}{Kwonsang Lee}, \bibinfo{person}{Falco~J
  Bargagli-Stoffi}, {and} \bibinfo{person}{Francesca Dominici}.}
  \bibinfo{year}{2020}\natexlab{}.
\newblock \showarticletitle{Causal rule ensemble: Interpretable inference of
  heterogeneous treatment effects}.
\newblock \bibinfo{journal}{\emph{arXiv preprint arXiv:2009.09036}}
  (\bibinfo{year}{2020}).
\newblock


\bibitem[\protect\citeauthoryear{Lemmerich, Atzmueller, and Puppe}{Lemmerich
  et~al\mbox{.}}{2016}]%
        {lemmerich2016fast}
\bibfield{author}{\bibinfo{person}{Florian Lemmerich}, \bibinfo{person}{Martin
  Atzmueller}, {and} \bibinfo{person}{Frank Puppe}.}
  \bibinfo{year}{2016}\natexlab{}.
\newblock \showarticletitle{Fast exhaustive subgroup discovery with numerical
  target concepts}.
\newblock \bibinfo{journal}{\emph{Data Mining and Knowledge Discovery}}
  \bibinfo{volume}{30}, \bibinfo{number}{3} (\bibinfo{year}{2016}),
  \bibinfo{pages}{711--762}.
\newblock


\bibitem[\protect\citeauthoryear{Letham, Rudin, McCormick, Madigan,
  et~al\mbox{.}}{Letham et~al\mbox{.}}{2015}]%
        {letham2015interpretable}
\bibfield{author}{\bibinfo{person}{Benjamin Letham}, \bibinfo{person}{Cynthia
  Rudin}, \bibinfo{person}{Tyler~H McCormick}, \bibinfo{person}{David Madigan},
  {et~al\mbox{.}}} \bibinfo{year}{2015}\natexlab{}.
\newblock \showarticletitle{Interpretable classifiers using rules and bayesian
  analysis: Building a better stroke prediction model}.
\newblock \bibinfo{journal}{\emph{The Annals of Applied Statistics}}
  \bibinfo{volume}{9}, \bibinfo{number}{3} (\bibinfo{year}{2015}),
  \bibinfo{pages}{1350--1371}.
\newblock


\bibitem[\protect\citeauthoryear{Li, Han, and Pei}{Li et~al\mbox{.}}{2001}]%
        {li2001cmar}
\bibfield{author}{\bibinfo{person}{Wenmin Li}, \bibinfo{person}{Jiawei Han},
  {and} \bibinfo{person}{Jian Pei}.} \bibinfo{year}{2001}\natexlab{}.
\newblock \showarticletitle{CMAR: Accurate and efficient classification based
  on multiple class-association rules}. In
  \bibinfo{booktitle}{\emph{Proceedings 2001 IEEE International Conference on
  Data Mining}}. IEEE, \bibinfo{pages}{369--376}.
\newblock


\bibitem[\protect\citeauthoryear{Lin, Zhong, Hu, Rudin, and Seltzer}{Lin
  et~al\mbox{.}}{2020}]%
        {LinEtAl20}
\bibfield{author}{\bibinfo{person}{Jimmy Lin}, \bibinfo{person}{Chudi Zhong},
  \bibinfo{person}{Diane Hu}, \bibinfo{person}{Cynthia Rudin}, {and}
  \bibinfo{person}{Margo Seltzer}.} \bibinfo{year}{2020}\natexlab{}.
\newblock \showarticletitle{Generalized and Scalable Optimal Sparse Decision
  Trees}. In \bibinfo{booktitle}{\emph{International Conference on Machine
  Learning {(ICML)}}}.
\newblock


\bibitem[\protect\citeauthoryear{Lipkovich, Dmitrienko, Denne, and
  Enas}{Lipkovich et~al\mbox{.}}{2011}]%
        {lipkovich_subgroup_2011}
\bibfield{author}{\bibinfo{person}{Ilya Lipkovich}, \bibinfo{person}{Alex
  Dmitrienko}, \bibinfo{person}{Jonathan Denne}, {and} \bibinfo{person}{Gregory
  Enas}.} \bibinfo{year}{2011}\natexlab{}.
\newblock \showarticletitle{Subgroup identification based on differential
  effect search—a recursive partitioning method for establishing response to
  treatment in patient subpopulations}.
\newblock \bibinfo{journal}{\emph{Statistics in Medicine}}
  \bibinfo{volume}{30}, \bibinfo{number}{21} (\bibinfo{year}{2011}),
  \bibinfo{pages}{2601--2621}.
\newblock


\bibitem[\protect\citeauthoryear{Malhotra, Craven, Ambrosius, Killeen, Haley,
  Cheung, Chonchol, Sarnak, Parikh, Shlipak, et~al\mbox{.}}{Malhotra
  et~al\mbox{.}}{2019}]%
        {malhotra2019effects}
\bibfield{author}{\bibinfo{person}{Rakesh Malhotra}, \bibinfo{person}{Timothy
  Craven}, \bibinfo{person}{Walter~T Ambrosius}, \bibinfo{person}{Anthony~A
  Killeen}, \bibinfo{person}{William~E Haley}, \bibinfo{person}{Alfred~K
  Cheung}, \bibinfo{person}{Michel Chonchol}, \bibinfo{person}{Mark Sarnak},
  \bibinfo{person}{Chirag~R Parikh}, \bibinfo{person}{Michael~G Shlipak},
  {et~al\mbox{.}}} \bibinfo{year}{2019}\natexlab{}.
\newblock \showarticletitle{Effects of intensive blood pressure lowering on
  kidney tubule injury in CKD: a longitudinal subgroup analysis in SPRINT}.
\newblock \bibinfo{journal}{\emph{American Journal of Kidney Diseases}}
  \bibinfo{volume}{73}, \bibinfo{number}{1} (\bibinfo{year}{2019}),
  \bibinfo{pages}{21--30}.
\newblock


\bibitem[\protect\citeauthoryear{McFowland~III, Somanchi, and
  Neill}{McFowland~III et~al\mbox{.}}{2018}]%
        {mcfowland2018efficient}
\bibfield{author}{\bibinfo{person}{Edward McFowland~III},
  \bibinfo{person}{Sriram Somanchi}, {and} \bibinfo{person}{Daniel~B Neill}.}
  \bibinfo{year}{2018}\natexlab{}.
\newblock \showarticletitle{Efficient discovery of heterogeneous treatment
  effects in randomized experiments via anomalous pattern detection}.
\newblock \bibinfo{journal}{\emph{arXiv preprint arXiv:1803.09159}}
  (\bibinfo{year}{2018}).
\newblock


\bibitem[\protect\citeauthoryear{Michalski, Carbonell, and Mitchell}{Michalski
  et~al\mbox{.}}{2013}]%
        {michalski2013machine}
\bibfield{author}{\bibinfo{person}{Ryszard~S Michalski},
  \bibinfo{person}{Jaime~G Carbonell}, {and} \bibinfo{person}{Tom~M Mitchell}.}
  \bibinfo{year}{2013}\natexlab{}.
\newblock \bibinfo{booktitle}{\emph{Machine learning: An artificial
  intelligence approach}}.
\newblock \bibinfo{publisher}{Springer Science \& Business Media}.
\newblock


\bibitem[\protect\citeauthoryear{Moodie, Chakraborty, and Kramer}{Moodie
  et~al\mbox{.}}{2012}]%
        {moodie2012q}
\bibfield{author}{\bibinfo{person}{Erica~EM Moodie}, \bibinfo{person}{Bibhas
  Chakraborty}, {and} \bibinfo{person}{Michael~S Kramer}.}
  \bibinfo{year}{2012}\natexlab{}.
\newblock \showarticletitle{Q-learning for estimating optimal dynamic treatment
  rules from observational data}.
\newblock \bibinfo{journal}{\emph{Canadian Journal of Statistics}}
  \bibinfo{volume}{40}, \bibinfo{number}{4} (\bibinfo{year}{2012}),
  \bibinfo{pages}{629--645}.
\newblock


\bibitem[\protect\citeauthoryear{Morucci, Orlandi, Roy, Rudin, and
  Volfovsky}{Morucci et~al\mbox{.}}{2020}]%
        {MorucciEtAl20}
\bibfield{author}{\bibinfo{person}{Marco Morucci}, \bibinfo{person}{Vittorio
  Orlandi}, \bibinfo{person}{Sudeepa Roy}, \bibinfo{person}{Cynthia Rudin},
  {and} \bibinfo{person}{Alexander Volfovsky}.}
  \bibinfo{year}{2020}\natexlab{}.
\newblock \showarticletitle{Adaptive Hyper-box Matching for Interpretable
  Individualized Treatment Effect Estimation}. In
  \bibinfo{booktitle}{\emph{Uncertainty in Artificial Intelligence}}.
\newblock


\bibitem[\protect\citeauthoryear{Nagpal, Wei, Vinzamuri, Shekhar, Berger, Das,
  and Varshney}{Nagpal et~al\mbox{.}}{2020}]%
        {nagpal2020interpretable}
\bibfield{author}{\bibinfo{person}{Chirag Nagpal}, \bibinfo{person}{Dennis
  Wei}, \bibinfo{person}{Bhanukiran Vinzamuri}, \bibinfo{person}{Monica
  Shekhar}, \bibinfo{person}{Sara~E Berger}, \bibinfo{person}{Subhro Das},
  {and} \bibinfo{person}{Kush~R Varshney}.} \bibinfo{year}{2020}\natexlab{}.
\newblock \showarticletitle{Interpretable subgroup discovery in treatment
  effect estimation with application to opioid prescribing guidelines}. In
  \bibinfo{booktitle}{\emph{Proceedings of the ACM Conference on Health,
  Inference, and Learning}}. \bibinfo{pages}{19--29}.
\newblock


\bibitem[\protect\citeauthoryear{Neill}{Neill}{2012}]%
        {neill2012fast}
\bibfield{author}{\bibinfo{person}{Daniel~B Neill}.}
  \bibinfo{year}{2012}\natexlab{}.
\newblock \showarticletitle{Fast subset scan for spatial pattern detection}.
\newblock \bibinfo{journal}{\emph{Journal of the Royal Statistical Society:
  Series B (Statistical Methodology)}} \bibinfo{volume}{74},
  \bibinfo{number}{2} (\bibinfo{year}{2012}), \bibinfo{pages}{337--360}.
\newblock


\bibitem[\protect\citeauthoryear{Novak, Lavra{\v{c}}, and Webb}{Novak
  et~al\mbox{.}}{2009}]%
        {novak2009supervised}
\bibfield{author}{\bibinfo{person}{Petra~Kralj Novak}, \bibinfo{person}{Nada
  Lavra{\v{c}}}, {and} \bibinfo{person}{Geoffrey~I Webb}.}
  \bibinfo{year}{2009}\natexlab{}.
\newblock \showarticletitle{Supervised descriptive rule discovery: A unifying
  survey of contrast set, emerging pattern and subgroup mining}.
\newblock \bibinfo{journal}{\emph{Journal of Machine Learning Research}}
  \bibinfo{volume}{10}, \bibinfo{number}{February} (\bibinfo{year}{2009}),
  \bibinfo{pages}{377--403}.
\newblock


\bibitem[\protect\citeauthoryear{Osofsky}{Osofsky}{1995}]%
        {osofsky1995effect}
\bibfield{author}{\bibinfo{person}{Joy~D Osofsky}.}
  \bibinfo{year}{1995}\natexlab{}.
\newblock \showarticletitle{The effect of exposure to violence on young
  children.}
\newblock \bibinfo{journal}{\emph{American Psychologist}} \bibinfo{volume}{50},
  \bibinfo{number}{9} (\bibinfo{year}{1995}), \bibinfo{pages}{782}.
\newblock


\bibitem[\protect\citeauthoryear{Pan, Wang, and Hara}{Pan
  et~al\mbox{.}}{2020}]%
        {pmlr-v108-pan20a}
\bibfield{author}{\bibinfo{person}{Danqing Pan}, \bibinfo{person}{Tong Wang},
  {and} \bibinfo{person}{Satoshi Hara}.} \bibinfo{year}{2020}\natexlab{}.
\newblock \showarticletitle{Interpretable Companions for Black-Box Models}. In
  \bibinfo{booktitle}{\emph{Artificial Intelligence and Statistics}},
  Vol.~\bibinfo{volume}{108}. \bibinfo{publisher}{PMLR},
  \bibinfo{address}{Online}, \bibinfo{pages}{2444--2454}.
\newblock


\bibitem[\protect\citeauthoryear{Rijnbeek and Kors}{Rijnbeek and Kors}{2010}]%
        {rijnbeek2010finding}
\bibfield{author}{\bibinfo{person}{Peter~R Rijnbeek} {and}
  \bibinfo{person}{Jan~A Kors}.} \bibinfo{year}{2010}\natexlab{}.
\newblock \showarticletitle{Finding a short and accurate decision rule in
  disjunctive normal form by exhaustive search}.
\newblock \bibinfo{journal}{\emph{Machine Learning}} \bibinfo{volume}{80},
  \bibinfo{number}{1} (\bibinfo{year}{2010}), \bibinfo{pages}{33--62}.
\newblock


\bibitem[\protect\citeauthoryear{Rothwell}{Rothwell}{2005}]%
        {rothwell2005subgroup}
\bibfield{author}{\bibinfo{person}{Peter~M Rothwell}.}
  \bibinfo{year}{2005}\natexlab{}.
\newblock \showarticletitle{Subgroup analysis in randomised controlled trials:
  importance, indications, and interpretation}.
\newblock \bibinfo{journal}{\emph{The Lancet}} \bibinfo{volume}{365},
  \bibinfo{number}{9454} (\bibinfo{year}{2005}), \bibinfo{pages}{176--186}.
\newblock


\bibitem[\protect\citeauthoryear{Rubin}{Rubin}{1974}]%
        {rubin1974estimating}
\bibfield{author}{\bibinfo{person}{Donald~B Rubin}.}
  \bibinfo{year}{1974}\natexlab{}.
\newblock \showarticletitle{Estimating causal effects of treatments in
  randomized and nonrandomized studies.}
\newblock \bibinfo{journal}{\emph{Journal of Educational Psychology}}
  \bibinfo{volume}{66}, \bibinfo{number}{5} (\bibinfo{year}{1974}),
  \bibinfo{pages}{688}.
\newblock


\bibitem[\protect\citeauthoryear{Sekhon}{Sekhon}{2011}]%
        {sekhon2011multivariate}
\bibfield{author}{\bibinfo{person}{Jasjeet~S Sekhon}.}
  \bibinfo{year}{2011}\natexlab{}.
\newblock \showarticletitle{Multivariate and propensity score matching software
  with automated balance optimization: the matching package for R}.
\newblock \bibinfo{journal}{\emph{Journal of Statistical Software}}
  \bibinfo{volume}{42} (\bibinfo{date}{May} \bibinfo{year}{2011}).
\newblock
Issue 7.


\bibitem[\protect\citeauthoryear{Solomon, Turunen, Ngandu, Peltonen,
  Lev{\"a}lahti, Helisalmi, Antikainen, B{\"a}ckman, H{\"a}nninen, Jula,
  et~al\mbox{.}}{Solomon et~al\mbox{.}}{2018}]%
        {solomon2018effect}
\bibfield{author}{\bibinfo{person}{Alina Solomon}, \bibinfo{person}{Heidi
  Turunen}, \bibinfo{person}{Tiia Ngandu}, \bibinfo{person}{Markku Peltonen},
  \bibinfo{person}{Esko Lev{\"a}lahti}, \bibinfo{person}{Seppo Helisalmi},
  \bibinfo{person}{Riitta Antikainen}, \bibinfo{person}{Lars B{\"a}ckman},
  \bibinfo{person}{Tuomo H{\"a}nninen}, \bibinfo{person}{Antti Jula},
  {et~al\mbox{.}}} \bibinfo{year}{2018}\natexlab{}.
\newblock \showarticletitle{Effect of the apolipoprotein E genotype on
  cognitive change during a multidomain lifestyle intervention: a subgroup
  analysis of a randomized clinical trial}.
\newblock \bibinfo{journal}{\emph{JAMA Neurology}} \bibinfo{volume}{75},
  \bibinfo{number}{4} (\bibinfo{year}{2018}), \bibinfo{pages}{462--470}.
\newblock


\bibitem[\protect\citeauthoryear{Su, Tsai, Wang, Nickerson, and Li}{Su
  et~al\mbox{.}}{2009}]%
        {su2009subgroup}
\bibfield{author}{\bibinfo{person}{Xiaogang Su}, \bibinfo{person}{Chih-Ling
  Tsai}, \bibinfo{person}{Hansheng Wang}, \bibinfo{person}{David~M Nickerson},
  {and} \bibinfo{person}{Bogong Li}.} \bibinfo{year}{2009}\natexlab{}.
\newblock \showarticletitle{Subgroup analysis via recursive partitioning}.
\newblock \bibinfo{journal}{\emph{Journal of Machine Learning Research}}
  \bibinfo{volume}{10}, \bibinfo{number}{Feb} (\bibinfo{year}{2009}),
  \bibinfo{pages}{141--158}.
\newblock


\bibitem[\protect\citeauthoryear{Wang and Rudin}{Wang and Rudin}{2015}]%
        {WangRu15}
\bibfield{author}{\bibinfo{person}{Fulton Wang} {and} \bibinfo{person}{Cynthia
  Rudin}.} \bibinfo{year}{2015}\natexlab{}.
\newblock \showarticletitle{Falling Rule Lists}. In
  \bibinfo{booktitle}{\emph{Artificial Intelligence and Statistics}}.
\newblock


\bibitem[\protect\citeauthoryear{Wang}{Wang}{2018}]%
        {wang2018multi}
\bibfield{author}{\bibinfo{person}{Tong Wang}.}
  \bibinfo{year}{2018}\natexlab{}.
\newblock \showarticletitle{Multi-value rule sets for interpretable
  classification with feature-efficient representations}. In
  \bibinfo{booktitle}{\emph{Advances in Neural Information Processing
  Systems}}. \bibinfo{pages}{10835--10845}.
\newblock


\bibitem[\protect\citeauthoryear{Wang and Lin}{Wang and Lin}{2019}]%
        {wang2019hybrid}
\bibfield{author}{\bibinfo{person}{Tong Wang} {and} \bibinfo{person}{Qihang
  Lin}.} \bibinfo{year}{2019}\natexlab{}.
\newblock \showarticletitle{Hybrid predictive model: When an interpretable
  model collaborates with a black-box model}.
\newblock \bibinfo{journal}{\emph{arXiv preprint arXiv:1905.04241}}
  (\bibinfo{year}{2019}).
\newblock


\bibitem[\protect\citeauthoryear{Wang, Morucci, Awan, Liu, Roy, Rudin, and
  Volfovsky}{Wang et~al\mbox{.}}{2021}]%
        {FLAME}
\bibfield{author}{\bibinfo{person}{Tianyu Wang}, \bibinfo{person}{Marco
  Morucci}, \bibinfo{person}{M.~Usaid Awan}, \bibinfo{person}{Yameng Liu},
  \bibinfo{person}{Sudeepa Roy}, \bibinfo{person}{Cynthia Rudin}, {and}
  \bibinfo{person}{Alexander Volfovsky}.} \bibinfo{year}{2021}\natexlab{}.
\newblock \showarticletitle{FLAME: A Fast Large-scale Almost Matching Exactly
  Approach to Causal Inference}.
\newblock \bibinfo{journal}{\emph{Journal of Machine Learning Research}}
  (\bibinfo{year}{2021}).
\newblock
\newblock
\shownote{Accepted, also arXiv 1707.06315.}


\bibitem[\protect\citeauthoryear{Wang, Rudin, Doshi, Liu, Klampfl, and
  MacNeille}{Wang et~al\mbox{.}}{2017}]%
        {wang2017bayesian}
\bibfield{author}{\bibinfo{person}{Tong Wang}, \bibinfo{person}{Cynthia Rudin},
  \bibinfo{person}{F Doshi}, \bibinfo{person}{Yimin Liu},
  \bibinfo{person}{Erica Klampfl}, {and} \bibinfo{person}{Perry MacNeille}.}
  \bibinfo{year}{2017}\natexlab{}.
\newblock \showarticletitle{A Bayesian Framework for Learning Rule Sets for
  Interpretable Classification}.
\newblock \bibinfo{journal}{\emph{Journal of Machine Learning Research}}
  \bibinfo{volume}{18}, \bibinfo{number}{70} (\bibinfo{year}{2017}),
  \bibinfo{pages}{1--37}.
\newblock


\bibitem[\protect\citeauthoryear{Wei, Dash, Gao, and Gunluk}{Wei
  et~al\mbox{.}}{2019}]%
        {wei2019generalized}
\bibfield{author}{\bibinfo{person}{Dennis Wei}, \bibinfo{person}{Sanjeeb Dash},
  \bibinfo{person}{Tian Gao}, {and} \bibinfo{person}{Oktay Gunluk}.}
  \bibinfo{year}{2019}\natexlab{}.
\newblock \showarticletitle{Generalized Linear Rule Models}. In
  \bibinfo{booktitle}{\emph{International Conference on Machine Learning}}.
  \bibinfo{pages}{6687--6696}.
\newblock


\bibitem[\protect\citeauthoryear{Yang, Rudin, and Seltzer}{Yang
  et~al\mbox{.}}{2017}]%
        {ynormalize_addang2016scalable}
\bibfield{author}{\bibinfo{person}{Hongyu Yang}, \bibinfo{person}{Cynthia
  Rudin}, {and} \bibinfo{person}{Margo Seltzer}.}
  \bibinfo{year}{2017}\natexlab{}.
\newblock \showarticletitle{Scalable Bayesian Rule Lists}. In
  \bibinfo{booktitle}{\emph{International Conference on Machine Learning}}.
\newblock


\bibitem[\protect\citeauthoryear{Yin and Han}{Yin and Han}{2003}]%
        {yin2003cpar}
\bibfield{author}{\bibinfo{person}{Xiaoxin Yin} {and} \bibinfo{person}{Jiawei
  Han}.} \bibinfo{year}{2003}\natexlab{}.
\newblock \showarticletitle{CPAR: Classification based on predictive
  association rules}. In \bibinfo{booktitle}{\emph{Proceedings of the 2003 SIAM
  International Conference on Data Mining}}. SIAM, \bibinfo{pages}{331--335}.
\newblock


\end{thebibliography}

\newpage
\section*{Appendix}
\subsection*{Proofs for Theorem 1 and Theorem 2}
\noindent\textbf{Proof} (of Theorem 1)
 We notate elements in $\mathbf{w}_{\bar{A}}$ as $\mathbf{w}_{\bar{A}} = \{\mathbf{v}_{\bar{A}},\gamma^{(0)}_{\bar{A}},\gamma^{(1)}_{\bar{A}},\gamma^{(2)}_{\bar{A}}\}$.
Since $\bar{A}(\mathbf{x}_i) = 1$ when $i \in I\backslash\mathcal{E}$, we rewrite $\bar{A}(\mathbf{x}_i)$ as
\begin{equation}
\bar{A}(\mathbf{x}_i) = T_iy_i + (1-T_i)(1-y_i).\label{eqn:proof_A}
\end{equation}
Expanding $\Theta(\bar{A},\mathbf{w}_{\bar{A}})$ using formula (4.10) and plugging in (\ref{eqn:proof_A}) yields
\begin{align}
& \log \Theta(\bar{A},\mathbf{w}_{\bar{A}})= \log p(\mathbf{w}_{\bar{A}}) +\sum_{y_i=1}\log \sigma\Big(\mathbf{v}_{\bar{A}}\mathbf{x}_i +\gamma^{(0)}_{\bar{A}}T_i+ \gamma^{(1)}_{\bar{A}} T_i+ \gamma^{(2)}_{\bar{A}} T_i \Big) \notag \\
&+\sum_{y_i=0}\log \Bigg[1- \sigma\Big( \mathbf{v}_{\bar{A}}\mathbf{x}_i+\gamma^{(0)}_{\bar{A}}T_i +\gamma^{(1)}_{\bar{A}} (1-T_i)\Big)\Bigg].
\end{align}
We then upper bound the conditional likelihood of data $\mathcal{D}$ and the prior of parameters given any rule set $A$ and $\mathbf{w}_A$.
\begin{align}
&\log\Theta(A,\mathbf{w}_A)= \sum_{i=1}^n\log P(y_i|\mathbf{x}_i,T_i;A,\mathbf{v}_A)+\log p(\mathbf{w}_A) = \log p(\mathbf{w}_A)\;\;+&   \notag \\
&\sum_{y_i = 1} \log \sigma\Big( \mathbf{v}_A\mathbf{x}_i + \gamma^{(0)}_AT_i + \gamma^{(1)}_A A(\mathbf{x}_i) + \gamma^{(2)}_A T_iA(\mathbf{x}_i) \Big) & (U_1)\notag \\
& + \sum_{y_i = 0} \log\Bigg[1-\sigma\Big( \mathbf{v}_A\mathbf{x}_i + \gamma^{(0)}_AT_i+\gamma^{(1)}_A A(\mathbf{x}_i)+\gamma^{(2)}_A T_i A(\mathbf{x}_i)\Big)\Bigg], &(U_2) \label{thm:up_BA}
\end{align}
where
\begin{align}
\gamma^{(1)}_A A(\mathbf{x}_i) + \gamma^{(2)}_A T_iA(\mathbf{x}_i) &\leq (\gamma^{(1)}_A + \gamma^{(2)}_A) T_iA(\mathbf{x}_i)\label{thm:up_1} \\
&\leq (\gamma^{(1)}_A + \gamma^{(2)}_A) T_i.\label{thm:up_2}
\end{align}
(\ref{thm:up_1}) follows since $\gamma^{(1)}\leq 0$ and $T_i\in\{0,1\}$, so $\gamma^{(1)}_A A(\mathbf{x}_i) \leq \gamma^{(1)}_A T_iA(\mathbf{x}_i) $. (\ref {thm:up_2}) follows because $\gamma^{(1)}+\gamma^{(2)}\geq 0$ and $A(\mathbf{x}_i)\in\{0,1\}$.
 Since $\sigma(x)$ increases monotonically with $x$, we get 
\begin{equation}\label{thm:up_3}
U_1\leq \sum_{y_i=1}\log \sigma\Big(\mathbf{v}_{A}\mathbf{x}_i +\gamma^{(0)}_{A}T_i+ \gamma^{(1)}_{A} T_i+ \gamma^{(2)}_{A} T_i \Big).
\end{equation}
Meanwhile, since $\gamma^{(2)}_A\geq-\gamma^{(1)}_A$,
\begin{align}
\gamma^{(1)}_A A(\mathbf{x}_i)+\gamma^{(2)}_A T_i A(\mathbf{x}_i) &\geq\gamma^{(1)}_A A(\mathbf{x}_i)-\gamma^{(1)}_A T_i A(\mathbf{x}_i)\notag \\
& = \gamma^{(1)}(1-T_i)A(\mathbf{x}_i) \geq  \gamma^{(1)}(1-T_i). \label{thm:up_4}
\end{align}
(\ref{thm:up_4}) follows because $\gamma^{(1)}\leq 0$ and $A(\mathbf{x})\in\{0,1\}$. Thus
\begin{equation}
U_2 \leq \sum_{y_i=0}\log \Bigg[1- \sigma\Big( \mathbf{v}_{A}\mathbf{x}_i+\gamma^{(0)}_{A}T_i +\gamma^{(1)}_{A} (1-T_i)\Big)\Bigg].\label{thm:up_5}
\end{equation}
Plugging (\ref{thm:up_3}) and (\ref{thm:up_5}) back into inequality (\ref{thm:up_BA}) yields
\begin{align}
\log\Theta(A,\mathbf{w}_A)&\leq  \log p(\mathbf{w}_A) + \sum_{y_i=1}\log \sigma\Big(\mathbf{v}_{A}\mathbf{x}_i +\gamma^{(0)}_{A}T_i+ \gamma^{(1)}_{A} T_i+ \gamma^{(2)}_{A} T_i \Big) \notag \\
& +\sum_{y_i=0}\log \Bigg[1- \sigma\Big( \mathbf{v}_{A}\mathbf{x}_i+\gamma^{(0)}_{A}T_i +\gamma^{(1)}_{A} (1-T_i)\Big)\Bigg] \notag\\
=&\log\Theta(\bar{A},\mathbf{w}_A)\leq \log\Theta(\bar{A},\mathbf{w}_{\bar{A}}). \label{eqn:proof_3}
\end{align}
(\ref{eqn:proof_3}) follows since $\mathbf{w}_{\bar{A}} = \max_{\mathbf{w}}\Theta(\bar{A},\mathbf{w})$.

\vspace{2mm}
\noindent\textbf{Proof} (Of Theorem 2)
Let $\emptyset$ denote an empty set where there are no rules and $\mathbf{w}_{\emptyset}$ is the optimal parameter corresponding to $\emptyset$. 
Since $\{A^*,\mathbf{w}_{A^*}\} \in \arg\min_{A,\mathbf{w}} F(A,\mathbf{w})$, then $F(A^*,\mathbf{w}_{A^*})\geq v^{[t]}$, i.e., 
\begin{equation}
\log \Theta(\mathcal{D};A^*,\mathbf{w}_{A^*}) + \log \text{Prior}(A^*) \geq v^{[t]}.\label{eqn:inequality}
\end{equation}
Now we find conditions where the above inequality always holds. We do it in the following two steps.
According to Theorem 1,  
\begin{equation}
\Theta(\mathcal{D};A^*,\mathbf{w}_{A^*}) \leq \Theta(\bar{A},\mathbf{w}_{\bar{A}}). 
 \label{eqn:2step1}
\end{equation}
The prior probability of selecting $A^*$ is
\begin{equation}
\text{Prior}(A^*) = \prod_l^L
\mathcal{L}_l h(M_l).\label{eqn:A2}
\end{equation}
We write out $h(M_{l})$ as the following, multiplying by 1 in disguise:
\begin{align}
h(M_l) = &\Gamma(\alpha_{l})\alpha_{l}\dots(\alpha_{l}+M_l-1)\times \frac{\Gamma(|\mathcal{A}_l|+\beta_{l}-M_l)(|\mathcal{A}_l|+\beta_{l}-M_l)\dots(|\mathcal{A}_l|+\beta_{l}-1)}{(|\mathcal{A}_l|+\beta_{l}-M_l)\dots(|\mathcal{A}_l|+\beta_{l}-1)}\notag \\
=&\frac{\Gamma(\alpha_{l})\Gamma(|\mathcal{A}_l|+\beta_{l})\alpha_{l}\dots(\alpha_{l}+M_l-1)}{(|\mathcal{A}_l|+\beta_{l}-M_l)\dots(|\mathcal{A}_l|+\beta_{l}-1)}\notag \\
\leq &\Gamma(\alpha_{l})\Gamma(|\mathcal{A}_l|+\beta_{l})\left(\frac{\alpha_{l}+m_l^{[t]}-1}{|\mathcal{A}_l|+\beta_{l}-1}\right)^{M_l}. \label{eqn:g1_last}
\end{align}
\normalsize
We also observe 
\begin{equation}\label{eqn:g2_0}
h(M_l) \leq h(0)
\end{equation}
for all $M_l$. To obtain this, we take the second derivative of $h(M_l)$ with respect to $M_l$, as follows:
\begin{align*}
h^{\prime\prime}(M_l) = &h(M_l)\Bigg(\left(\sum_{k=1}^{\infty}\frac{1}{k+|\mathcal{A}_l|+\beta_l-M_l-1}-\frac{1}{k+M_l+\alpha_l-1}\right)^2+ \\ 
&\left( \sum_{k=1}^{\infty}\frac{1}{k+|\mathcal{A}_l|+\beta_l-M_l-1}\right)^2 + \left(\sum_{k=1}^{\infty}\frac{1}{k+M_l+\alpha_l-1} \right)^2\Bigg)\\
&>0.
\end{align*}
Therefore, $h(M_l)$ is strictly convex, and $h(M_l) \leq \max\{h(0),g(|\mathcal{A}_l|)\}=h(0)$. 
Combining (\ref{eqn:A2}) with (\ref{eqn:g1_last}) and (\ref{eqn:g2_0}) we have
\begin{align}\label{eqn:1_A_llh_bound}
\text{Prior}(A^*) & =\mathcal{L}_{l^\prime}h(M_{l^\prime}) \prod_{l=1,...L,l\neq l^\prime} \mathcal{L}_{l}h(M_{l})\notag \\
&\leq \mathcal{L}_{l^\prime} \Gamma(\alpha_{l})\Gamma(|\mathcal{A}_l|+\beta_{l})\left(\frac{|\mathcal{A}_l|+\alpha_l-1}{|\mathcal{A}_l|+\beta_{l}-1}\right)^{M_l} \prod_{l=1,...L,l\neq l^\prime} \mathcal{L}_{l}h(0) \notag \\
& = \left(\frac{M_l+\alpha_{l^\prime}-1}{|\mathcal{A}_l|+\beta_{l^\prime}-1}\right)^{M_{l^\prime}} \cdot \prod_{l}^L \mathcal{L}_l h(0) \notag \\
&=
\left(\frac{M_l+\alpha_{l^\prime}-1}{|\mathcal{A}_l|+\beta_{l^\prime}-1}\right)^{M_{l^\prime}}\cdot p(\emptyset), 
\end{align}
which means
\begin{equation}\label{eqn:Aempty_diff}
\log \text{Prior}(A^*) \leq \log p(\emptyset) +M_{l^\prime} \log \left( \frac{M_l+\alpha_{l^\prime}-1}{|\mathcal{A}_l|+\beta_{l^\prime}-1}\right).
\end{equation}
Now we apply inequality (\ref{eqn:inequality}) and substituting in (\ref{eqn:2step1}) and (\ref{eqn:1_A_llh_bound}), we find
\begin{align}\label{eqn:objective_cmp}
\Theta(\bar{A},\mathbf{w}_{\bar{A}}) + \log p(\emptyset) +M_{l^\prime} \log \left( \frac{M_l+\alpha_{l^\prime}-1}{|\mathcal{A}_l|+\beta_{l^\prime}-1}\right) \geq v^{[t]}.
\end{align}
Since $\alpha_{l^\prime} < \beta_{l^\prime}$, so $\log \left( \frac{|M_l|+\alpha_{l^\prime}-1}{|\mathcal{A}_l|+\beta_{l^\prime}-1}\right) <0$, therefore 
\begin{equation}
m_l^{[t]}=\frac{\log\Theta(\bar{A},\mathbf{w}_{\bar{A}})+\log p(\emptyset)-v^{[t]} }{\log \left(\frac{|\mathcal{A}_l|+\beta_{l}-1}{m_l^{[t-1]}+\alpha_{l}-1}\right)}    
\end{equation}
which holds for each $l \in \{1,2...,L\}$. Thus, the total number of rules in $A^*$ is bounded by
\begin{equation}
|A^*|\leq \sum_{l=1}^L   m_l^{[t]},
\end{equation}
which completes the proof.
\subsection*{More Examples of CRS Models}
More examples are provided in Tables \ref{tab:crowdfundinga}, \ref{tab:crowdfundingb}, and \ref{tab:crowdfundingc}.
\begin{table}[ht]
\centering
\caption{A CRS model learned from the in-vehicle recommender system data. The support is 28.6\% and the average treatment effect is 0.43, evaluated on the test set}\label{tab:crowdfundinga}
\begin{tabular}{llc}
\toprule
        & \multicolumn{1}{c}{\textbf{Rules}}         \\\hline
\textbf{If} &  age $\geq 41$ \textbf{\emph{and}} destination $=$ no immediate destination \textbf{\emph{and}} expiration $=$ 1 day   \\
& \textbf{\emph{OR}} destination = work \textbf{\emph{and}} degree $\geq$ college  \\
\textbf{Then}        &  the instance is in the subgroup  \\ 
\textbf{Else}    &the instance is not in the subgroup      \\ \bottomrule                
\end{tabular}\vspace{-2mm}
\end{table}

\begin{table}[ht]
\centering
\footnotesize
\caption{A CRS model learned from the juvenile delinquency data. The support is 24.0\% and the average treatment effect is 0.26, evaluated on the test set}\label{tab:crowdfundingb}
\begin{tabular}{llc}
\toprule
        & \multicolumn{1}{c}{\textbf{Rules}}         \\\hline
\textbf{If} &  has anyone -- including family members or friends -- attacked you without a weapon $\neq$ ``yes''\\
& \textbf{\emph{and}} did your friends steal something worth less than \$5 = 
``at least some of them'' \\
&\textbf{\emph{and}} did your friends steal something worth more than \$50 $\neq$ unknown   \\
& \textbf{\emph{OR}} did your friends steal something worth more than \$50 $\neq$ ``no'' \\
& \textbf{\emph{and}} did your friends use marijuana or hashish = ``at least some of them ''\\
&\textbf{\emph{and}} did your friends hit or threaten to hit someone without any reason = ``at least very few of them \\
& \textbf{\emph{OR}} has anyone -- including family members or friends -- ever attacked you with a weapon $\neq$ ``yes \\
& \textbf{\emph{and}} did your friends use marijuana or hashish = ``at least some of them ''\\
&\textbf{\emph{OR}} did your friends steal something worth more than \$50 $\neq$ ``no \\
& \textbf{\emph{and}} did your friends hit or threaten to hit someone without any reason = ``at least some of them'' \\
& \textbf{\emph{and}} did your friends get drunk once in a while = ``at least most of them'' \\
\textbf{Then}        &  the instance is in the subgroup  \\ 
\textbf{Else}    &the instance is not in the subgroup      \\ \bottomrule                
\end{tabular}\vspace{-2mm}
\end{table}

\begin{table}[ht]
\centering
\caption{A CRS model learned from the young voter turnout dataset. The support is 24.3\% and the average treatment effect is 0.045, evaluated on the test set}\label{tab:crowdfundingc}
\begin{tabular}{llc}
\toprule
        & \multicolumn{1}{c}{\textbf{Rules}}         \\\hline
\textbf{If} &  congress district = 13 \textbf{\emph{and}} party = republican  \\
\textbf{Then}        &  the instance is in the subgroup  \\ 
\textbf{Else}    &the instance is not in the subgroup      \\ \bottomrule                
\end{tabular}\vspace{-2mm}
\end{table}
\end{document}